%% file: arxiv.tex
\RequirePackage{letltxmacro}
\LetLtxMacro{\LaTeXtextbf}{\textbf}

\documentclass[conference]{IEEEtran}
\IEEEoverridecommandlockouts

\LetLtxMacro{\textbf}{\LaTeXtextbf}

\usepackage{cite}
\usepackage{amsmath,amssymb,amsfonts}
\usepackage{graphicx}
\usepackage{textcomp}

\usepackage{CJKutf8} 
\usepackage[inline]{enumitem} 
\usepackage{booktabs} 
\usepackage{url}
\usepackage{fancyvrb} 
\usepackage{float} 
\usepackage{adjustbox} 
\usepackage{xcolor} 
\usepackage{multirow} 
\usepackage{colortbl} 
\usepackage{vcell} 
\usepackage[ruled, vlined, linesnumbered, commentsnumbered]{algorithm2e} 
\usepackage{tabularx} 
\usepackage{tikz} 
\usetikzlibrary{svg.path} 
\usepackage{scalerel} 

\definecolor{darkgreen}{RGB}{6,128,4}
\definecolor{pistachio}{RGB}{104,119,33} 
\definecolor{bordeaux}{RGB}{186,33,33}
\newcommand{\xmltag}[1]{\textcolor{darkgreen}{\textbf{#1}}}
\newcommand{\xmlattr}[1]{\textcolor{pistachio}{\textbf{#1}}}
\newcommand{\xmlvalue}[1]{\textcolor{bordeaux}{\textbf{#1}}}

\newcommand{\wwlatinchar}{%
\unskip\tikz[baseline=(current bounding box.south)]{%
\draw[red, line width=0.2pt] (0,0) -- (0.6em,0);%
}\ignorespaces%
}
\newcommand{\bwpinyin}{%
\unskip\tikz[baseline=(current bounding box.south)]{%
\draw[red, line width=0.2pt] (0,0) -- (1em,0);%
\draw[red, line width=0.2pt] (0.5em,0) -- (0.5em,0.5em);%
}\ignorespaces%
}
\newcommand{\wwpinyinsyll}{%
\unskip\tikz[baseline=(current bounding box.south)]{%
\draw[red, line width=0.2pt] (0,0) -- (0.5em,0.2em);%
\draw[red, line width=0.2pt] (0.5em,0.2em) -- (1em,0);%
}\ignorespaces%
}
\newcommand{\imeconf}{%
\unskip\tikz[baseline=-0.5ex]{%
  \draw[blue, -latex] (0.9em,0.25ex) -- (0.2em,0.25ex);%
  \draw[blue, -latex] (0.9em,-0.25ex) -- (0.2em,-0.25ex);%
}\ignorespaces%
}
\newcommand{\bwchinese}{%
\unskip\tikz[baseline=-0.5ex]{%
  \draw[blue, line width=0.2pt] (0,-0.35ex) -- (0,0.35ex);%
  \draw[blue, ->] (0,0) -- (1em,0);%
}\ignorespaces%
}

\SetKwProg{Fn}{Function}{:}{end}
\SetKwFunction{GetText}{GetText}
\SetKwComment{Comment}{$\triangleright$\ }{}
\SetAlFnt{\small}
\SetAlCapNameFnt{\small}
\SetAlCapFnt{\small}
\AtBeginEnvironment{algorithm}{
	\DontPrintSemicolon
}

\SetFuncSty{textsc}
\SetCommentSty{commentfont}
\SetKwProg{Fn}{Function}{}{}
\SetKwInOut{Input}{input}
\SetKwInOut{Output}{output}
\SetKwComment{Comment}{$\triangleright$\ }{}

\def\BibTeX{{\rm B\kern-.05em{\sc i\kern-.025em b}\kern-.08em
T\kern-.1667em\lower.7ex\hbox{E}\kern-.125emX}}

\newcommand\submittedtext{%

  \footnotesize This is the accepted version of the article published in IEEE Access: 
  R. Crotti, G. Denaro, Z. Du and R. M. Martín, "Hylog: A Hybrid Approach to Logging Text Production in Non-alphabetic Scripts," in IEEE Access, doi: 10.1109/ACCESS.2026.3718565.}

\newcommand\submittednotice{%
\begin{tikzpicture}[remember picture,overlay]
\node[anchor=south,yshift=10pt] at (current page.south) {\fbox{\parbox{\dimexpr0.65\textwidth-\fboxsep-\fboxrule\relax}{\submittedtext}}};
\end{tikzpicture}%
}

\usepackage[hidelinks]{hyperref} 

\begin{document}

\title{Hylog: A Hybrid Approach to Logging Text Production in Non-alphabetic Scripts\\
\thanks{\textsuperscript{*}Equal contribution.\\This work was supported by co-financing from the Italian National Research Plan PRIN 2022, within the EU initiative Next Generation EU, awarded to Ricardo Muñoz [2022EYX28N]. All authors are members of the research team associated with this project.}
}

\author{\IEEEauthorblockN{Roberto Crotti*}
\IEEEauthorblockA{\textit{Department of Informatics, Systems and Communication} \\
\textit{University of Milano-Bicocca}\\
Milan, Italy \\
roberto.crotti@unimib.it \\
ORCID: 0009-0009-5634-5765}
\and
\IEEEauthorblockN{Giovanni Denaro*}
\IEEEauthorblockA{\textit{Department of Informatics, Systems and Communication} \\
\textit{University of Milano-Bicocca}\\
Milan, Italy \\
giovanni.denaro@unimib.it \\
ORCID: 0000-0002-7566-8051}
\and
\IEEEauthorblockN{Zhiqiang Du*}
\IEEEauthorblockA{\textit{Department of Interpreting and Translation} \\
\textit{Alma Mater Studiorum University of Bologna}\\
Bologna, Italy \\
zhiqiang.du2@unibo.it \\
ORCID: 0000-0002-6659-250X}
\and
\IEEEauthorblockN{Ricardo Mu\~noz Martín*}
\IEEEauthorblockA{\textit{Department of Interpreting and Translation} \\
\textit{Alma Mater Studiorum University of Bologna}\\
Bologna, Italy \\
ricardo.munoz@unibo.it \\
ORCID: 0000-0001-6049-9673}
}

\maketitle

\renewcommand\fbox{\fcolorbox{red}{white}}
\setlength{\fboxrule}{2pt} 
\submittednotice

\newif\ifrev
\newenvironment{rev}{%
  \ifrev
    \color{red}%
  \fi
}{}

\newif\ifrevv
\newenvironment{revv}{%
  \ifrevv
    \color{orange}%
  \fi
}{}

\begin{abstract}
Research keyloggers are essential for cognitive studies of text production, yet most fail to capture the on-screen transformations performed by Input Method Editors (IMEs) for non-alphabetic scripts. To address this methodological gap, we present Hylog, a novel hybrid logging system that combines analytical keylogging with ecological text logging for a more complete and finer-grained analysis. Our modular, open-source system uses plug-ins for standard applications (Microsoft Word, Google Chrome) to capture both keyboard output and rendered text, which a hybridizer module then synchronizes into a dual trace. To validate the system's technical feasibility and demonstrate its analytical capabilities, we conducted a proof-of-concept study where two volunteers translated a text into simplified Chinese. Hylog successfully captured keypresses and temporal intervals between Latin letters, Chinese characters, and IME confirmations---some measurements invisible to traditional keyloggers. The resulting data enable the formulation of new, testable hypotheses about the cognitive restrictions and affordances at different linguistic layers in IME-mediated typing. Our plug-in architecture enables extension to other IME systems and fosters more inclusive multilingual text-production research.
\end{abstract}


 \input{1-introduction}
 \input{2-logging-non-alphabetic-scripts}
 \input{3-hybrid-keylogging}
 \input{4-empirical-testing}
 \input{5-discussion-related-work-conclusions}

\bibliographystyle{IEEEtran}
\bibliography{bibliography.bib}

\newpage

\begin{IEEEbiography}[{\includegraphics[width=1in,height=1.25in,clip,keepaspectratio]{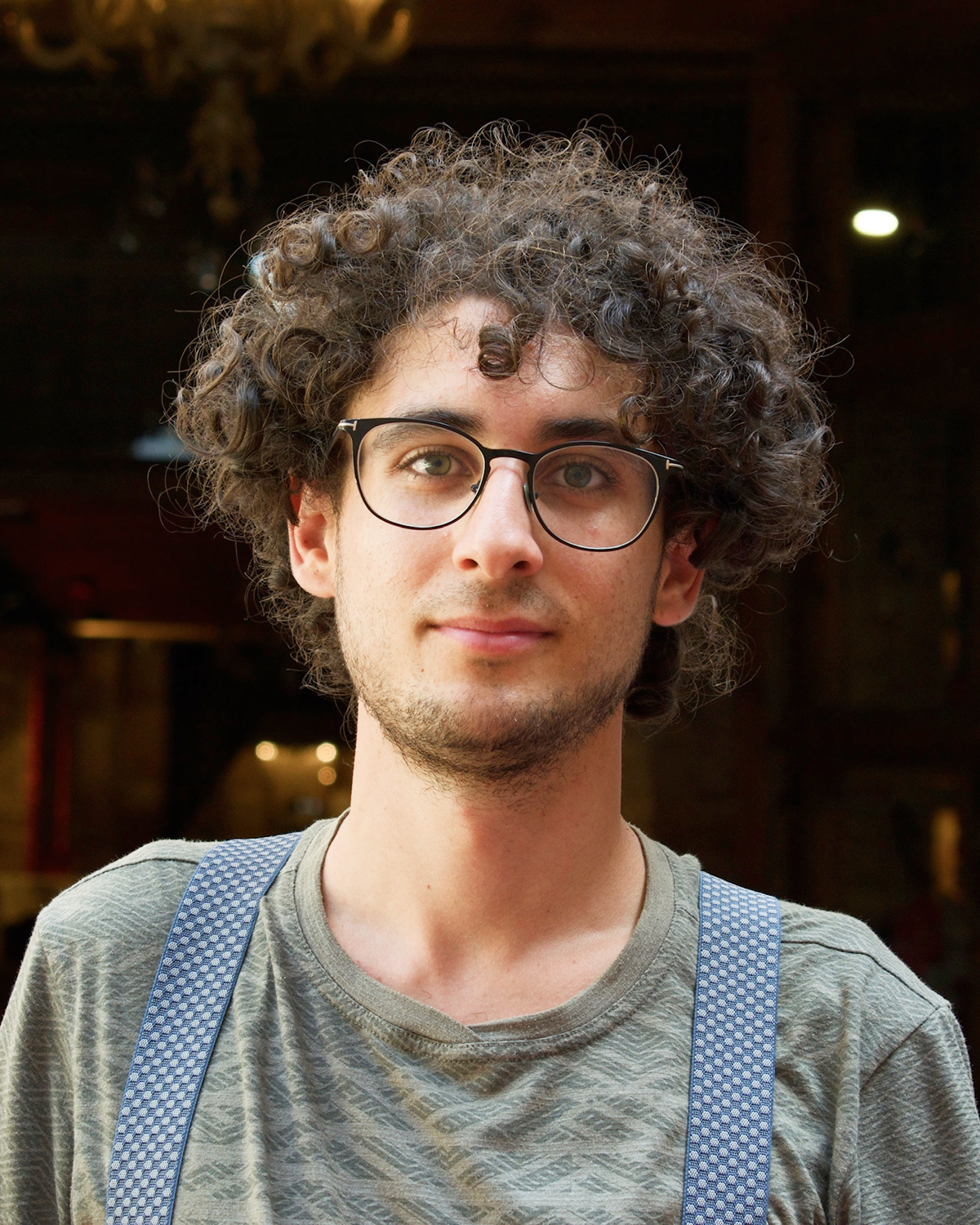}}]{Roberto Crotti} worked as a research fellow at the University of Milano-Bicocca, where he contributed to the PRIN project “Big Sistah – Quantifying the wellbeing of multilingual remote workers in real-time”, focusing on the design of multimodal logging systems for complex input methods (IME). 

He holds a Bachelor’s degree in Computer Science and a Master’s degree in Theory and Technology of Communication. His research interests lie at the intersection of computer science, human–computer interaction and cognitive and social psychology, with particular focus on the design and ethical implications of interactive systems. 

His education combines computer science with studies in communication and human-centered design, providing an interdisciplinary perspective on interactive systems.
\end{IEEEbiography}

\vskip 0pt plus -1fil

\begin{IEEEbiography}[{\includegraphics[width=1in,height=1.25in,clip,keepaspectratio]{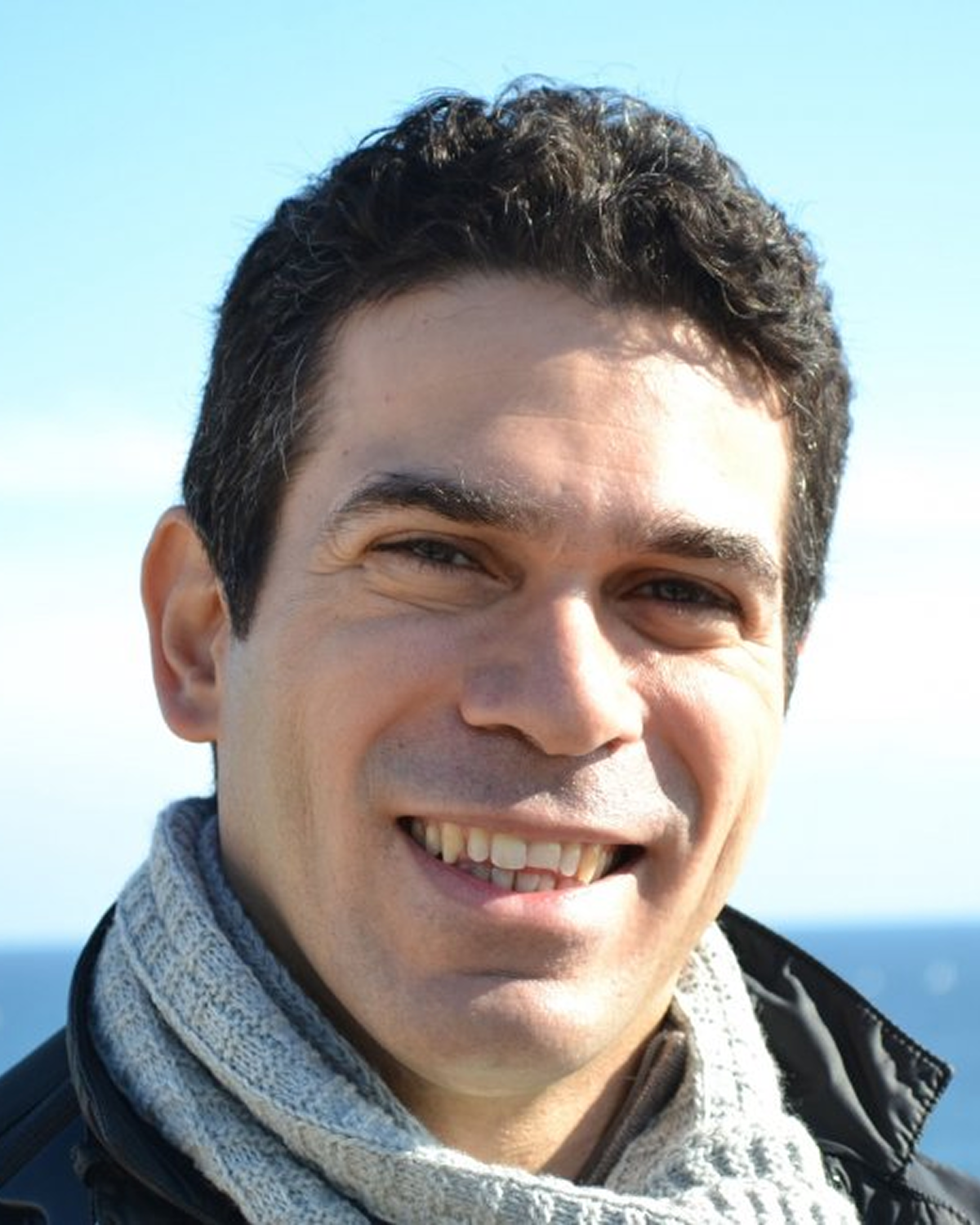}}]{Giovanni Denaro} (Member, IEEE) is an Associate Professor at the University of Milano-Bicocca, where his research focuses on software engineering, with particular emphasis on software testing and analysis. 

He received his Master’s degree and PhD in Computer Engineering from Politecnico di Milano and has conducted research at several international academic and industrial institutions, including CEFRIEL, University College London, and Università della Svizzera Italiana. He has been involved in multiple national and European research projects.

He is Associate Editor of IEEE Transactions on Software Engineering and of the Software Quality Journal, and also served in leading roles and as a program committee member of ICSE, FSE, ASE, and ISSTA.
\end{IEEEbiography}

\vskip 0pt plus -1fil

\begin{IEEEbiography}[{\includegraphics[width=1in,height=1.25in,clip,keepaspectratio]{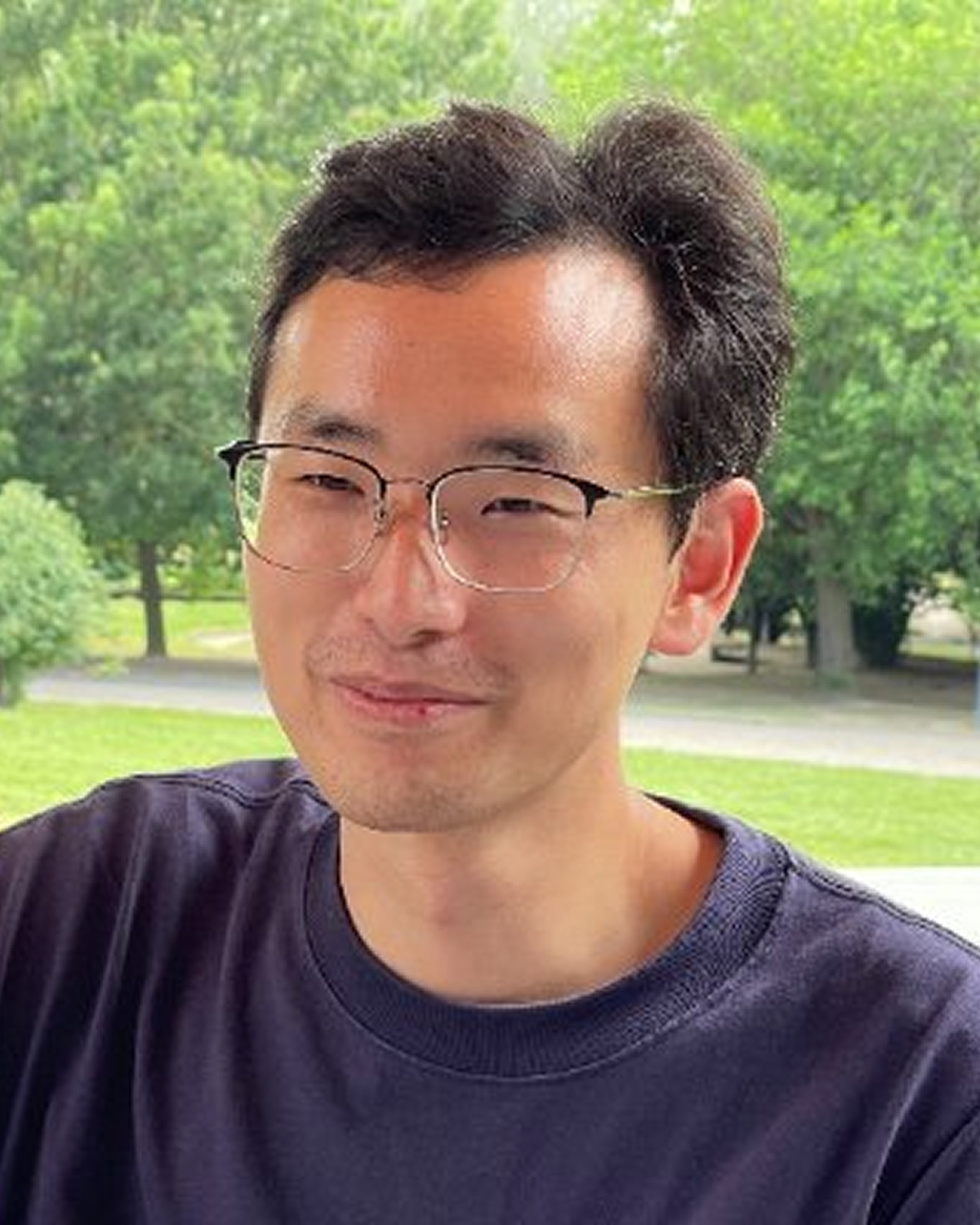}}]{Zhiqiang Du} is a PhD fellow at the University of Bologna, where he completed his PhD in 2024. His dissertation examined Chinese interpreting trainees’ performance using computer-assisted interpreting (CAI) tools in remote settings from a cognitive perspective. 

His current research interests include the cognitive-situated study of interpreting, CAI tool use and assessment, remote simultaneous interpreting, keylogging, and information-seeking behavior. His work employs multimodal data collection methods in oral tasks, including keylogging and audio and screen recording. He focuses on understanding of interpreters’ cognitive processes and performance in multilectal mediated communication environments. 
\end{IEEEbiography}

\vskip 0pt plus -1fil

\begin{IEEEbiography}[{\includegraphics[width=1in,height=1.25in,clip,keepaspectratio]{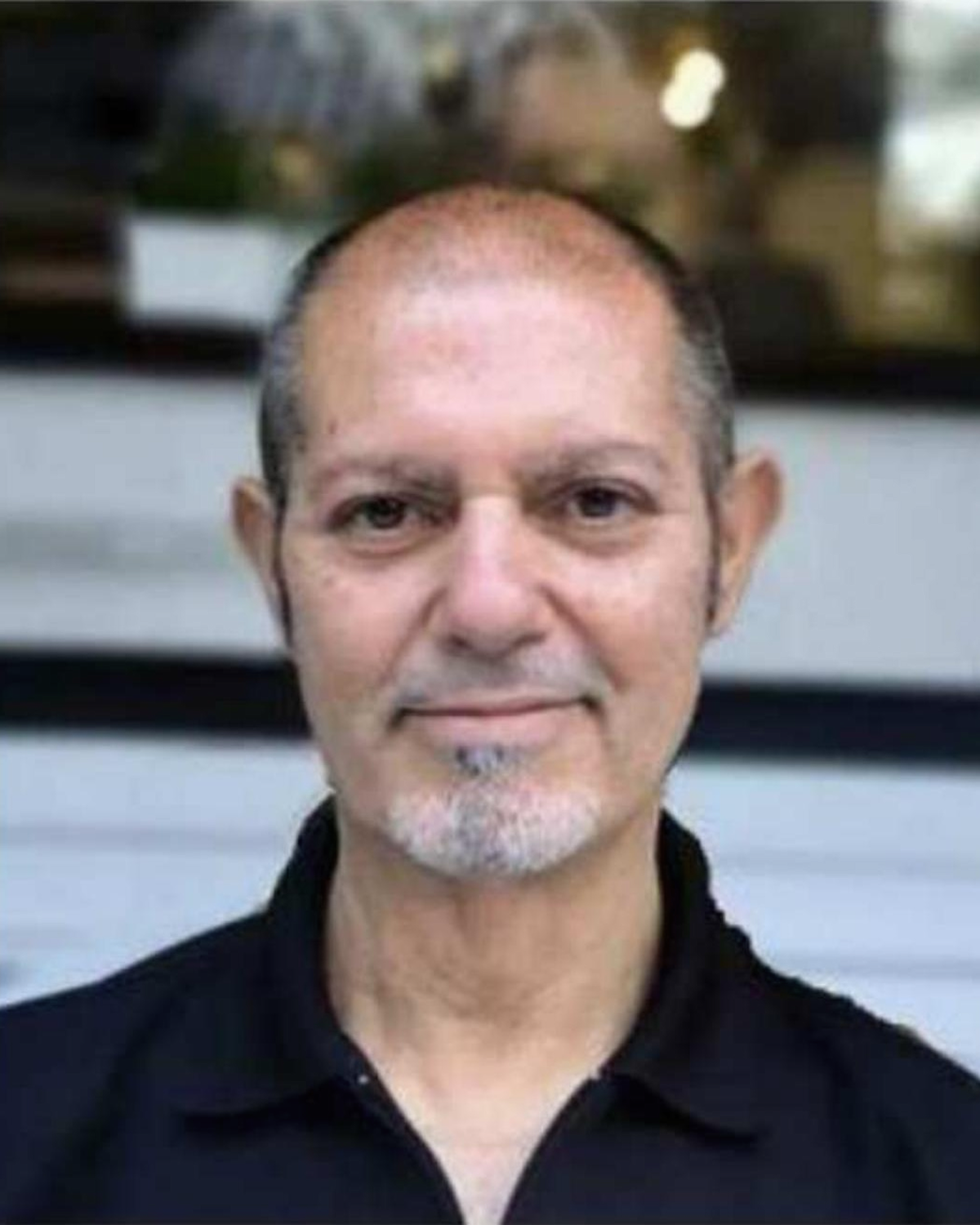}}]{Ricardo Mu\~noz Martín} is a Full Professor at the University of Bologna, where his research focuses on cognitive translation and interpreting studies, particularly the development of cognitive translatology and empirical methods to trace translation processes such as keylogging. 

He received his PhD in Hispanic Languages \& Literatures from the University of California, Berkeley, and holds degrees in Translation \& Interpreting and Anglo-Germanic Philology from Spanish universities. He leads the Laboratory for Multilectal Mediated Communication \& Cognition (MC2 Lab) and has published extensively on cognitive approaches to mediated communication and translation studies. He also serves on editorial boards of several international journals and has substantial experience in research leadership and academic service.
\end{IEEEbiography}
\end{document}

%% file: 1-introduction.tex
\section{Introduction}\label{introduction}
The generalized digitalization of text-production processes has settled keyloggers as core tools to study the cognitive processes involved. 
Keyloggers yield detailed recordings of human--computer interactions during text production through computer-based input mechanisms\begin{rev}. Researchers then\end{rev} infer various aspects of the underlying cognitive processes from those recordings~\cite{couto2017}.
Keyloggers enable researchers to examine how keystroke latencies, typing speed, and revisions relate to text quality in both L1 and L2 writing \cite{vandermeulen2024} and expertise development \cite{xu2014}. Integration with other tools, such as screen recording \cite{raido2023, rosa2018}, further captures writers' interactions with auxiliary tools such as web browsers and dictionaries \cite{leijten2019}.
The analysis of \textit{interkeystroke-intervals,} or IKIs---the time span or latency between releasing a key and pressing the next one---has also been applied to research on translation processes \cite{lacruz2012, munoz2018} and in biometric studies. 

Keyloggers offer two critical advantages: temporal precision and unobtrusiveness \cite{Munoz2025}. Temporal precision refers to millisecond-level timestamps that enable measurement of typos, corrections, and intervals that may signal diverse mental states and processes.
Unobtrusiveness ensures ecological validity by invisibly monitoring keyboard and mouse activities during text production\begin{rev}. This avoids\end{rev} the interference associated with think-aloud protocols, cued retrospection, and human observation \cite{van2009}.
However, not all applications that register typists' interactions are strictly keyloggers. We use \textit{input loggers} as an umbrella term to refer to all research applications that register the typists' text production as they interact with computers. Input loggers can be classified as either \textit{keyloggers} or \textit{text loggers}, and offer different degrees of temporal precision and unobtrusiveness. To the best of our knowledge, five applications have been used for academic research \begin{rev}{on text-production}\end{rev}~\cite{wengelin2023}: CyWrite \cite{cywrite}, GenoGraphiX-Log \cite{genographixlog}, Inputlog \cite{leijten2013}, ScriptLog \cite{scriptlog}, and Translog II \cite{carl2012}.

\textit{Text loggers}, like Translog II, record on-screen text output and can capture any visual event. However, they miss keyboard events that do not manifest visually, such as those produced by modifier keys and other non-printing keys. Text loggers generally have lower temporal precision than keyloggers because operating system software layers can introduce unpredictable delays in how input events are rendered on screen. \textit{Keyloggers} such as GenoGraphiX-Log, CyWrite, Scriptlog, and Inputlog record keyboard outputs directly through low-level operating system hooks. Low-level integration with the operating system enables these tools to record keystroke events with very accurate temporal precision.

Nevertheless, except for Inputlog, these keyloggers require typists to enter text within a built-in text area with a rudimentary internal text processor.
They can thus be described as \textit{non-ecological} \begin{rev}because they force the user to type into constrained interfaces. Unlike Inputlog, they do not allow users to work with common text processors and tools.\end{rev}
Such loggers are rarely deemed unobtrusive, which is why researchers have largely abandoned them. At the time of this research, we could not locate a downloadable release of ScriptLog. Similarly, the current release of CyWrite on GitHub is just demonstrative and can hardly support empirical research. Consequently, Inputlog has become the most widely adopted keylogger in text-production studies.

\begin{rev}
Other custom input loggers have been incorporated into tools for  biometric authentication based on the distinctive typing dynamics of users~\cite{Sun2016, Murphy2017,Vural2014}, 
or tools for evaluating the usability of user interfaces based on the interactions from users~\cite{Katerina2018},  
or yet tools for identifying emotional states based on keystroke dynamics~\cite{Epp2011}. 
These input loggers are not distributed as standalone tools, and are generally designed to tightly (and ecologically) integrate in the techniques for which they were developed. Overall they are all keyloggers, working similarly to Inputlog.
\end{rev}

We consider the general lack of appropriate tools a crucial limitation, as most research findings rest on experiments using alphabetic languages, with limited attention to quantitative analyses of text-production processes in non-alphabetic scripts. This limitation reflects structural differences: entering text in Chinese differs fundamentally from alphabetic typing in ways that determine what keyloggers can capture and analyze. Several features of Chinese and the way it is typed in illustrate why.

First, Chinese is a tonal language, which means that different words with the same sounds have different meanings and characters depending on their intonation. Second, Chinese words vary considerably in length.
Most everyday words have two characters whose meanings blend\begin{rev}. Often, the second character acts as the head and \end{rev} the first as a modifier.
However, technical vocabulary often has three, four, or more characters\begin{rev}. In addition,\end{rev} four-character expressions like \textit{chengyu} idioms \begin{CJK}{UTF8}{gbsn}(成语)\end{CJK} are common in writing.
Term length depends on how the underlying concept breaks down into morphemes.\footnote{In alphabetic languages, words combine minimal \textit{meaningful} components, called \textit{morphemes}, such as roots and suffixes, as opposed to syllables, which do not necessarily mean anything by themselves. For instance, in \textit{visible}, \textit{invisible}, and \textit{visibility}, \textit{vis} is the root or stem, while \textit{-ible}, \textit{in-}, and \textit{-ity} are affixes. In Chinese, each character sounds like a syllable but carries meaning and roughly corresponds to a morpheme. In a way, then, Chinese words combine components of more balanced semantic weight.} Third, alphabetic languages use spaces to separate words but Chinese does not. Both humans and computers need to identify or infer word boundaries instead. Tone-blind transcription, morphological variability and lack of word boundaries make segmentation quite challenging.

Many current input methods for non-alphabetic scripts do not link physical keystrokes directly to characters. Instead, non-alphabetic strings result from interposed applications known as \textit{Input Method Editors} (IMEs), which mediate between the alphabetic (typically \texttt{QWERTY}) keyboard and the symbols of the target script. In Chinese, for example, typists typically use the \textit{pinyin} ('spell-sound') input method: they type Roman alphabet letters (Latin letters) to enter phonetic transcriptions of Chinese characters. However, in pinyin input, tones are ignored. This means that a single pinyin transcription can match many different characters.

A Chinese IME configured within the writing software (e.g., Microsoft Word) dynamically presents sets of candidate characters matching the typed pinyin transcription. These characters are listed by decreasing likelihood, mainly based on frequency.
The typist then chooses one character, typically by pressing a number key\begin{rev}. For example, 1 (or \texttt{SPACEBAR}) chooses the first candidate, 2 the second, and so on.\end{rev}
The Latin letters used for the phonetic transcription on screen are then replaced with the chosen Chinese character(s). Machine learning features improve IME results on a computer over time. Thus, pressing \texttt{SPACEBAR} as the default choice is very common. Besides, typists may choose to keep typing the transcription of several characters at once, often those forming a word or ready-made expression, and only enter the rendering by the end.

Studies of text-production processes in non-alphabetic scripts \cite{wang2001, dasilva2017, lu2021} are significantly fewer than those on alphabetic languages \cite{lu2020}. Furthermore, very few employ keyloggers for quantitative analysis.
One example is \cite{dasilva2017} which combines Translog II with eyetracking to examine the cognitive processes involved in translating and post-editing between Chinese and Portuguese. 
In that study, the metrics are based on eyetracking data---such as gaze fixations---and linked to the Chinese text produced. The text is extracted from Translog II output and tokenized using an automatic segmenter. While the findings are of interest, the main technology is eyetracking, and no measures are defined or used that rely on keyboard input data. In another study focusing on revisions while writing in Chinese, researchers had to manually extract the frequencies of the phenomena of interest \cite{lu2021}. Manual approaches of this kind require considerable, often unacceptable effort, and arguably foster underpowered quantitative studies with limited samples.

This paper introduces Hylog, a novel hybrid system for logging non-alphabetic scripts that extends Inputlog's capabilities and makes the following primary contributions: 
\begin{enumerate*}[label=(\roman*)]
    \item a novel, ecological text-logging architecture featuring a Dynamic Snapshot Window (DSW) algorithm for efficient logging in standard text processors;
    \item a hybridizer module that robustly synchronizes low-level keystroke data with high-level rendered text output from IMEs;
    \item an open-source implementation of this system, provided as a resource to the research community;
    \item a proof-of-concept validation demonstrating the system's ability to capture fine-grained temporal metrics in Chinese, thereby enabling new avenues for quantitative research.
\end{enumerate*}

This combined approach enables our logging technique to operate in an ecological fashion while registering both keyboard outputs and their corresponding screen renderings. 

We tested the effectiveness of our approach in a small study where two experienced volunteers translated a heating pad instruction leaflet from English into simplified Chinese. We addressed the challenge of using keylogger results to analyze IKIs, traditionally defined and studied in text-production activities with alphabetic languages \cite{miller2000, wengelin2006, baaijen2012, barkaoui2019, vandermeulen2024, ivaska2025}, and tested whether our hybrid technique enables precise measurements.
\begin{rev}We compared the measurements produced by our hybrid approach with those obtained using only Inputlog 9.5 as a baseline. The results provide empirical evidence that our measurements\end{rev} are significantly more complete and precise.

The remainder of this paper is organized as follows. \S~\ref{section:logging-non-alphabetic-scripts} introduces and illustrates the challenges for extending the analysis of IKIs (our case study) to text-production activities with non-alphabetic scripts. It also discusses the limitations of current keylogging approaches with respect to those challenges. \S~\ref{sec:hylog} describes the hybrid, ecological keylogging approach that we introduce in this paper, in order to effectively address keylogging for non-alphabetic scripts. \S~\ref{sec:system-validation} summarizes a case study where we evaluated the effectiveness of our approach.\begin{rev}\S~\ref{sec:hylog:for:behavioral:analysis} shows the possible use of Hylog for behavioral analysis, presenting the kinds of insights researchers can derive.\end{rev} \S~\ref{sec:conclusions} discusses the relationships between our work and the new version of Inputlog for Chinese, summarizes our contributions, and highlights possible research directions.

%% file: 2-logging-non-alphabetic-scripts.tex
\section{Logging non-alphabetic scripts}
\label{section:logging-non-alphabetic-scripts}

This section explores the limitations of current input loggers in capturing text-production phenomena with non-alphabetic scripts, particularly IKIs. We focus on two widely used tools—the keylogger Inputlog and the text logger Translog II—and discuss their performance in sample writing sessions in English and Chinese. The examples underscore their limitations in capturing fine-grained text-production dynamics in simplified Chinese. Additionally, our examples reveal the challenges of applying standard IKI analysis techniques to non-alphabetic input systems.
\begin{rev} The last part of this section also summarizes and compares the capabilities and limitations of all the input loggers we considered in the introduction.\end{rev}

\subsection{Keyloggers-Inputlog}\label{subsection:inputlog-analysis}

Inputlog timestamps every keystroke, computes the corresponding IKIs, identifies the number of words in the sentence, and classifies each IKI into categories including \textit{within-word} and \textit{between-word}. This distinction matters because typists using alphabetical languages typically enter full words in one go, producing shorter within-word IKIs---deviations from planned behavioral sequences that may mirror salient environmental distractors, higher cognitive effort, mental fatigue, or stress---and longer between-word IKIs, which may host planning or assessment activities.

To illustrate Inputlog's capabilities, one author typed Laozi's popular saying both in Chinese \begin{CJK}{UTF8}{gbsn}`千里之行，始于足下'\end{CJK} and its English translation, `A journey of a thousand miles begins with a single step', in controlled sessions. Figure \ref{fig:inputlog-english-example} illustrates a sample record from a session entering the English version: the letters typed by the author appear in red and IKIs are shown in square brackets: black numbers mark within-word IKIs and blue numbers mark between-word IKIs, also separated with double underscores (\textit{\_\_}) for legibility:

\begin{figure*}[h]
 \begin{flushleft}
 \small 
    \noindent\texttt{\textcolor{red}A\textcolor{blue}{\_\_[1206]\_\_}\textcolor{red}j[368]\textcolor{red}o[278]\textcolor{red}u[284]\textcolor{red}r[301]\textcolor{red}n[243]\textcolor{red}e[234]\textcolor{red}y\textcolor{blue}{\_\_[421]\_\_}%
    \textcolor{red}o[291]\textcolor{red}f\textcolor{blue}{\_\_[505]\_\_}\textcolor{red}a\textcolor{blue}{\_\_[562]\_\_}\\\textcolor{red}t[310]\textcolor{red}h[251]\textcolor{red}o[236]\textcolor{red}u[278]\textcolor{red}s[266]\textcolor{red}a[350]\textcolor{red}{nd}\textcolor{blue}{\_\_[465]\_\_}%
    \textcolor{red}m[208]\textcolor{red}i[201]\textcolor{red}{le}[231]\textcolor{red}s\textcolor{blue}{\_\_[420]\_\_}\\\textcolor{red}{be}[224]\textcolor{red}{gins}\textcolor{blue}{\_\_[448]\_\_}
    \textcolor{red}{wi}[265]\textcolor{red}t[219]\textcolor{red}h\textcolor{blue}{\_\_[409]\_\_}\textcolor{red}a\textcolor{blue}{\_\_[422]\_\_}\\\textcolor{red}s[238]\textcolor{red}i[210]\textcolor{red}n[287]\textcolor{red}g[257]\textcolor{red}l[211]\textcolor{red}e\textcolor{blue}{\_\_[704]\_\_}\textcolor{red}s[231]\textcolor{red}t[228]\textcolor{red}e[204]\textcolor{red}p}
\end{flushleft}
\caption{Example of Inputlog's record of English typing.}
\label{fig:inputlog-english-example}
\end{figure*}

Table~\ref{tab:pauses:eng} reports the statistics that Inputlog provides for the input in Figure \ref{fig:inputlog-english-example}. The results align well with expectations: the number of between-word IKIs matches the number of white spaces in the sentence, and the number of within-word IKIs is consistent with the structure of that sentence. Overall, Inputlog reliably captures low-level typing features in English, enabling detailed temporal analyses.\footnote{A minor discrepancy appeared: Inputlog reported 26 within-word IKIs, whereas our plug-in (described in \S~\ref{sec:hylog}) identified 27. The likely cause was a misclassification in the analyzer module, which mislabeled the final IKI between \textit{e} and \textit{p} as an end-of-sentence interval. Yet, the logging itself remained accurate; the problem lay in post-processing.}

\begin{table}
\centering
\caption{\textbf{Interkeystroke interval statistics as reported by Inputlog for our English-writing sample session.}}
\begin{tabular}{lccc}
\toprule
& \textbf{Count} & \textbf{Mean (ms)} & \textbf{Sdev (ms)}\\
\midrule
\textbf{Between words} & 10 & 556 & 233 \\ 
\textbf{Within words} & 26 & 258 & 42 \\ 
\bottomrule
\end{tabular}
\label{tab:pauses:eng}
\end{table}

Consider now the version in Chinese, \begin{CJK}{UTF8}{gbsn}`千里之行，始于足下'\end{CJK}. The author types in the pinyin transcription of \begin{CJK}{UTF8}{gbsn}`千里\textit{(qiānlǐ)}之\textit{(zhī)}行\textit{(xíng)}'\end{CJK}, then presses `1' to select the first IME candidate, replacing the transcription with those Chinese characters. He continues with a Latin comma (interpreted as Chinese comma) and types \textit{shiyuzuxia}---for \begin{CJK}{UTF8}{gbsn}`始\textit{(shǐ)}于\textit{(yú)}足下\textit{(zúxià)}'\end{CJK}---again accepting the first candidate to complete the sentence \begin{CJK}{UTF8}{gbsn}`始于足下'\end{CJK}.

In short, typists enter Latin letters and read both these letters and Chinese characters on the screen. Hence, for temporal analyses we can differentiate at least five IKI subcategories in pinyin typing, depending on whether they occur between

\begin{enumerate}[label=(\alph*)]
\item two Latin letters within a pinyin syllable
\item two pinyin syllables (or Chinese characters, once entered)
\item two Chinese words
\item any event and an IME confirmation (before IME conversion)
\item an IME confirmation and any event (after IME conversion)
\end{enumerate}

IKIs between Latin letters (category (a) in the list above) are computed only within each pinyin syllable; IKIs of boundary letters at transitions between syllables (e.g., between e--s in \textit{xue'sheng} \begin{CJK}{UTF8}{gbsn}`学生'\end{CJK}) are excluded from this category and computed as IKIs in the next higher category (category (b)), between pinyin Latin syllables or Chinese characters (since each syllable typically becomes one character once converted by the IME). Likewise, IKIs between characters are computed only within words, while transitions between words are treated as between-word IKIs (category (c)). 
Yet, IKIs before the characters used to confirm IME candidates (category (d)) and transitions between words that occur after an IME conversion (category (e)) are computed separately, due to the specificity of those events. 
These types of IKIs may flag different cognitive activities (e.g., assessment, planning, reading, problem solving, etc).

A robust input logger for research should identify these types of IKIs, but Inputlog 9.5---designed for alphabetical languages---does not capture the Chinese characters inserted via the IME editor, nor does it recognize the structure of the pinyin input (versions to work with Korean and Chinese are currently being tested).\footnote{\url{www.inputlog.net}} Consequently, Inputlog 9.5 and earlier versions do not categorize IKIs specific to pinyin input (e.g., intervals between syllables and IME confirmations). It misidentifies word boundaries and thus fails to detect between-word IKIs, leading to systematic errors in IKI computation.

\begin{table*}[h]
\centering
\caption{Chinese example---keydown--keydown intervals reported by Inputlog (I) and the hybrid keylogger Hylog (H).}
\begin{tabular}{llcccccc}
\toprule
& & \multicolumn{2}{c}{\textbf{Num. pauses}} & \multicolumn{2}{c}{\textbf{IKI mean (ms)}} & \multicolumn{2}{c}{\textbf{IKI SD (ms)}} \\
\cline{3-8}
& & \textbf{I} & \textbf{H} & \textbf{I} & \textbf{H} & \textbf{I} & \textbf{H} \\
\midrule
\textbf{Latin letters} & $\wwlatinchar$ & 22 & 16 & 1387 & 1245 & 1052 & 941 \\
\textbf{Pinyin syllables} & $\wwpinyinsyll$ & n.a. & 2 & n.a. & 1159 & n.a. & 301 \\
\textbf{Chinese words} & $\bwpinyin$ & 0 & 5 & 0 & 1303 & 0 & 112 \\
\textbf{IME, before} & $\imeconf$ & n.a. & 2 & n.a. & 3180 & n.a. & 1305 \\
\textbf{IME, after} & $\bwchinese$ & n.a. & 1 & n.a. & 1124 & n.a. & 0 \\
\bottomrule
\end{tabular}
\label{tab:pauses:chinese}
\end{table*}

Table~\ref{tab:pauses:chinese} compares the results obtained for Chinese typing by Inputlog 9.5 for alphabetical scripts with those computed adding Hylog (columns labeled \textit{H}). Hylog is the novel keylogger for non-alphabetic languages that we introduce in this paper, and that we present in detail in \S~\ref{sec:hylog}. The first column also assigns a distinct symbol to each IKI type that are used below to illustrate where each IKI occurred during the writing session. Red color is used to mark IKIs between linguistic units (letters, syllables, words), while blue indicates behavioral IKIs, i.e., latencies before and after the typist confirms an IME candidate.

\vspace{1em}\noindent
{
\small
\texttt{%
q\wwlatinchar{}i\wwlatinchar{}a\wwlatinchar{}n%
\wwpinyinsyll{}l\wwlatinchar{}i\bwpinyin{}z%
\wwlatinchar{}h\wwlatinchar{}i\bwpinyin{}x%
\wwlatinchar{}i\wwlatinchar{}n\wwlatinchar{}g%
\imeconf{}1\bwchinese{}\bwpinyin{},\wwlatinchar{}\\s%
\wwlatinchar{}h\wwlatinchar{}i\bwpinyin{}y%
\wwlatinchar{}u\bwpinyin{}z%
\wwlatinchar{}u\wwpinyinsyll{}x%
\wwlatinchar{}i\wwlatinchar{}a%
\imeconf{}1}
}
\vspace{1em}

Inputlog 9.5 reports 22 IKIs between Latin letters and punctuation marks and none between words because it processes the entire pinyin string as a single uninterrupted sequence of letters rather than as several words. Yet, the sentence consists of six words, with 16 \textit{between-letter} IKIs and 5 \textit{between-word} IKIs, which distorts the corresponding mean and standard deviation values. These examples show that Inputlog  performs well with alphabetic languages but does not yet process non-alphabetic scripts such as Chinese, where an IME mediates the keyboard input through several layers of conversion.

Pinyin input entails at least two structural layers absent in alphabetic languages. First, phonetic composition occurs at the syllable level (addressed in \S~\ref{sec:language-unit-based-analysis}). Second, as illustrated in the example above, characters combine into words. The conversion from Latin letters to Chinese characters through IMEs adds a selection–verification subtask that anchors temporal and cognitive phenomena to IME operations---something absent from alphabetic typing---that need to be accounted for.

\subsection{Text loggers - Translog II}
\label{subsection:translog-analysis}

The second approach, \textit{text logging}, captures the textual output as it appears on screen rather than the keystrokes themselves. Translog II is a text logger used in many translation studies \cite{dragsted2013, sjrup2013, screen2016, heilmann2021, sun2021, qassem2023, sinulingga2023, qassem2024, wang02092024, wang2025}. Text loggers have clear advantages for Chinese: they record both the pinyin input in Roman alphabet and the resulting Chinese characters as rendered on screen. Figure~\ref{lst:translog-pinyin-session} presents an excerpt from the log that Translog II produces for the same Chinese example. It captures the typing of Latin letters and the events for IME choices, with complete information on both the pinyin syllables and the Chinese characters that replace them. 

\begin{CJK}{UTF8}{gbsn}

\begin{figure}[ht]

\caption{Sample log collected with Translog II.}
\label{lst:translog-pinyin-session}
\noindent\rule{\linewidth}{0.5pt}
\vspace{0em}

\begin{adjustbox}{max width=\linewidth}
\begin{BVerbatim}[fontsize=\small, commandchars=\\\{\}, numbers=left, stepnumber=1, numbersep=5pt, xleftmargin=1em]
\xmltag{<Events>}
  \xmltag{<System} \xmlattr{Time}=\xmlvalue{"0"} \xmlattr{Value}=\xmlvalue{"START"} \xmltag{/>}
  [...]
  \xmltag{<Key} \xmlattr{Time}=\xmlvalue{"13125"} \xmlattr{Cursor}=\xmlvalue{"0"} \xmlattr{Type}=\xmlvalue{"IME"} \xmlattr{Value}=\xmlvalue{"[X]"} \xmltag{/>}
  \xmltag{<Key} \xmlattr{Time}=\xmlvalue{"13484"} \xmlattr{Cursor}=\xmlvalue{"0"} \xmlattr{Type}=\xmlvalue{"IME"} \xmlattr{Value}=\xmlvalue{"[I]"} \xmltag{/>}
  \xmltag{<Key} \xmlattr{Time}=\xmlvalue{"13891"} \xmlattr{Cursor}=\xmlvalue{"0"} \xmlattr{Type}=\xmlvalue{"IME"} \xmlattr{Value}=\xmlvalue{"[N]"} \xmltag{/>}
  \xmltag{<Key} \xmlattr{Time}=\xmlvalue{"14297"} \xmlattr{Cursor}=\xmlvalue{"0"} \xmlattr{Type}=\xmlvalue{"IME"} \xmlattr{Value}=\xmlvalue{"[G]"} \xmltag{/>}
  \xmltag{<Key} \xmlattr{Time}=\xmlvalue{"18609"} \xmlattr{Cursor}=\xmlvalue{"0"} \xmlattr{IMEtext}=\xmlvalue{"QIANLIZHIXING"} 
    → \xmlattr{Type}=\xmlvalue{"insert"} \xmlattr{Value}=\xmlvalue{"千里之行"} \xmltag{/>}
  [...]
  \xmltag{<Key} \xmlattr{Time}=\xmlvalue{"26266"} \xmlattr{Cursor}=\xmlvalue{"0"} \xmlattr{Type}=\xmlvalue{"IME"} \xmlattr{Value}=\xmlvalue{"[I]"} \xmltag{/>}
  \xmltag{<Key} \xmlattr{Time}=\xmlvalue{"26687"} \xmlattr{Cursor}=\xmlvalue{"0"} \xmlattr{Type}=\xmlvalue{"IME"} \xmlattr{Value}=\xmlvalue{"[A]"} \xmltag{/>}
  \xmltag{<Key} \xmlattr{Time}=\xmlvalue{"27297"} \xmlattr{Cursor}=\xmlvalue{"5"} \xmlattr{IMEtext}=\xmlvalue{"SHIYUZUXIA"} 
    → \xmlattr{Type}=\xmlvalue{"insert"} \xmlattr{Value}=\xmlvalue{"始于足下"} \xmltag{/>}
  \xmltag{<System} \xmlattr{Time}=\xmlvalue{"33672"} \xmlattr{Value}=\xmlvalue{"STOP"} \xmltag{/>}
\xmltag{</Events>}
\end{BVerbatim}
\end{adjustbox}
\noindent\rule{\linewidth}{0.5pt}
\end{figure}

\end{CJK}

Text loggers excel at capturing complex editing operations like copy-pasting or non-contiguous text selection. However, they record screen-rendered events rather than physical keystrokes, introducing timing noise and delays, since the keystrokes must be processed by both the operating system and the word processing applications before they appear in the output eventually.
This raises doubts about the reliability of text loggers' timing data and, by extension, the validity of interpretations derived from it.

Furthermore, text loggers cannot capture several fine-grained aspects of writing that keyloggers can. We have already introduced \textit{interkey latency} or interkeystroke interval (IKI), the lapse between the keyup of one key and the keydown of the next. Other keystroke metrics include \textit{dwell time} (or \textit{hold latency})---the duration between keydown and keyup---and \textit{press latency} (keydown--keydown intervals), commonly used when higher precision is unnecessary or unavailable. Biometric studies often measure keyup-to-keyup intervals \textit{(release latency)} as well \cite{teh2013, yang2021, acien2022, kasprowski2022}. Despite terminological variation, these metrics capture closely related temporal phenomena.

Text-production studies, whether in writing or translation, also address the differences between dwell time and IKIs \textit{(flight time).} For instance, \cite{munoz2021} report that IKIs tend to follow the pattern of their interspersed dwell times but are more sensitive to environmental and situational variables. \textit{Rollover}---where a second key starts to be pressed before the first is completely released---is a salient phenomenon \cite{Dhakal2018} but still largely understudied. Such metrics cannot be extracted from text-logger recordings, whereas keyloggers can capture them directly. Inputlog 9.5, for example, computes \textit{press latency} (keydown--keydown) in its pause analysis, but it also records both keydown and keyup events in its general analysis, so dwell times, IKIs, and rollovers can also be derived.\footnote{Some newer tools, such as \textit{FlexKeyLogger} \cite{TianCushing2025}, also rely on Inputlog's built-in pause-analysis to process keystroke data and thus reproduce the same methodological problem: they compute inter-keystroke intervals as keydown–keydown latencies, merging dwell and flight times into a single measure.}

Finally, all current research text loggers, including Translog II, are non-ecological: they can only record visible text entered within their own interface, and fail to capture other subtasks, such as online searching or cross-application typing. These limitations compromise both the precision and the ecological validity required for comprehensive research of naturalistic behavior. The suitability of Inputlog, Translog II, or alternatives such as Pynput (e.g., \cite{du2024})  ultimately depends on the research question. However, the technical challenges involved often exceed the expertise of many researchers—a gap that motivated the project presented in the following section.

\begin{rev}
\subsection{Comparison}
Table \ref{tab:loggers:comparison}
analytically compares the capabilities and limitations of Inputlog~9.5, Translog~II, and the other input loggers that we mentioned in the introduction, that is, GenoGraphiX-Log, CyWrite, ScriptLog and the tool-custom, non standalone keyloggers that we surveyed from literature beyond the research on text production. The last row of the table refers to Hylog, the input logger that we introduce in this paper and that we present in detail in the next sections. 
For each tool, the table reports  descriptive data in the first two columns, that is, the name of the tool (column \textit{Input logger}), the seminal paper(s), and whether it is a keylogger or a text logger (column \textit{Type}). Our input logger Hylog hybridizes keylogging and  text logging, aiming to synergistically exploit the benefits of both approaches, while also contrasting the respective limitations. 

\begin{table*}[]
\centering
\begin{rev}
\caption{Comparison of input loggers.}
\begin{tabular}{ll|ccc} 
\toprule
\textbf{Input logger}            & \textbf{Type}                                                                                             & \textbf{Ecological}             & \begin{tabular}[c]{@{}c@{}}\textbf{Non alphabetic}\\\textbf{support}\end{tabular} & \begin{tabular}[c]{@{}c@{}}\textbf{Timestamps for }\\\textbf{keyup and keydown}\end{tabular}  \\ 
\midrule
\textbf{Inputlog 9.5~\cite{leijten2013}}            &     Keylogger                                                                                                                                     & \textcolor[rgb]{0,0.502,0}{Yes} & \textcolor{red}{No}                                                               & \textcolor[rgb]{0,0.502,0}{Yes}                                                               \\
\textbf{Translog II\cite{carl2012}}             & Text logger                                                                                                    & \textcolor{red}{No}             & \textcolor[rgb]{0,0.502,0}{Yes}                                                   & \textcolor{red}{No}                                                                           \\
\textbf{GenoGraphiX-Log~\cite{genographixlog}}        & Keylogger                                                                                                         & \textcolor{red}{No}             & \textcolor{red}{No}                                                               & \textcolor[rgb]{0,0.502,0}{Yes}                                                               \\
\textbf{CyWrite\cite{cywrite}}                 &   Keylogger                                                                                                                                           & \textcolor{red}{No}             & \textcolor{red}{No}                                                               & \textcolor[rgb]{0,0.502,0}{Yes}                                                               \\
\textbf{ScriptLog~\cite{wengelin2023}}            &     Keylogger                                                                                                                                        & \textcolor{red}{No} & \textcolor{red}{No}                                                               & \textcolor[rgb]{0,0.502,0}{Yes}                                                               \\
\textbf{Tool-custom keyloggers~\cite{Sun2016,Murphy2017,Vural2014,Epp2011,Katerina2018}} &           Keylogger                  & \textcolor[rgb]{0,0.502,0}{Yes} & \textcolor{red}{No}                                                               & \textcolor[rgb]{0,0.502,0}{Yes}                                                               \\
\textbf{Hylog}                   & Hybrid                                                                                                           & \textcolor[rgb]{0,0.502,0}{Yes} & \textcolor[rgb]{0,0.502,0}{Yes}                                                   & \textcolor[rgb]{0,0.502,0}{Yes}                                                               \\
\bottomrule
\end{tabular}
\label{tab:loggers:comparison}
\end{rev}
\end{table*}

The data reported in the table derive from our direct experience with Inputlog~9.5, Translog~II, GenoGraphiX-Log and CyWrite.  
In the case of ScripLog, as we were unable to identify any available release of the tool, we refer to the information reported by Wengelin and Johansson~\cite{wengelin2023}. Similarly, as the keloggers used in other domains are not distributed a standalone tools, we refer to the information from the corresponding seminal papers~\cite{Sun2016,Murphy2017,Vural2014,Epp2011,Katerina2018}. 

We compared the input loggers across three dimensions: whether or not they are ecological (Table~\ref{tab:loggers:comparison}, third column), if they include support for non-alphabetic languages (fourth column), and their ability to record fine-grained timestamps for keyup and keydown events (fifth column).

In the table, we observe that no input logger performs ideally in all dimensions. 
On one hand, Translog II, the only text logger, is the only tool that supports non-alphabetic writing, but it is not ecological and suffers of imprecise timestamps as it does not have visibility of keyup and keydown events, and incurs delays from the operating system. On the hand, all keyloggers, can provide fine-grained keystroke timestamps and part of them are indeed ecological, but none supports non-alphabetic writing, as they cannot identify logograms and symbols introduced by IMEs. 

These gaps shaped our design of Hylog, a hybrid key-and-text logging approach that combines the strengths of both text loggers and keyloggers, as we discuss in the next sections of the paper.
\end{rev}

%% file: 3-hybrid-keylogging.tex
\section{Hybrid keylogging}\label{sec:hylog}
\begin{figure*}[h]
\centering
\includegraphics[width=0.7\textwidth]{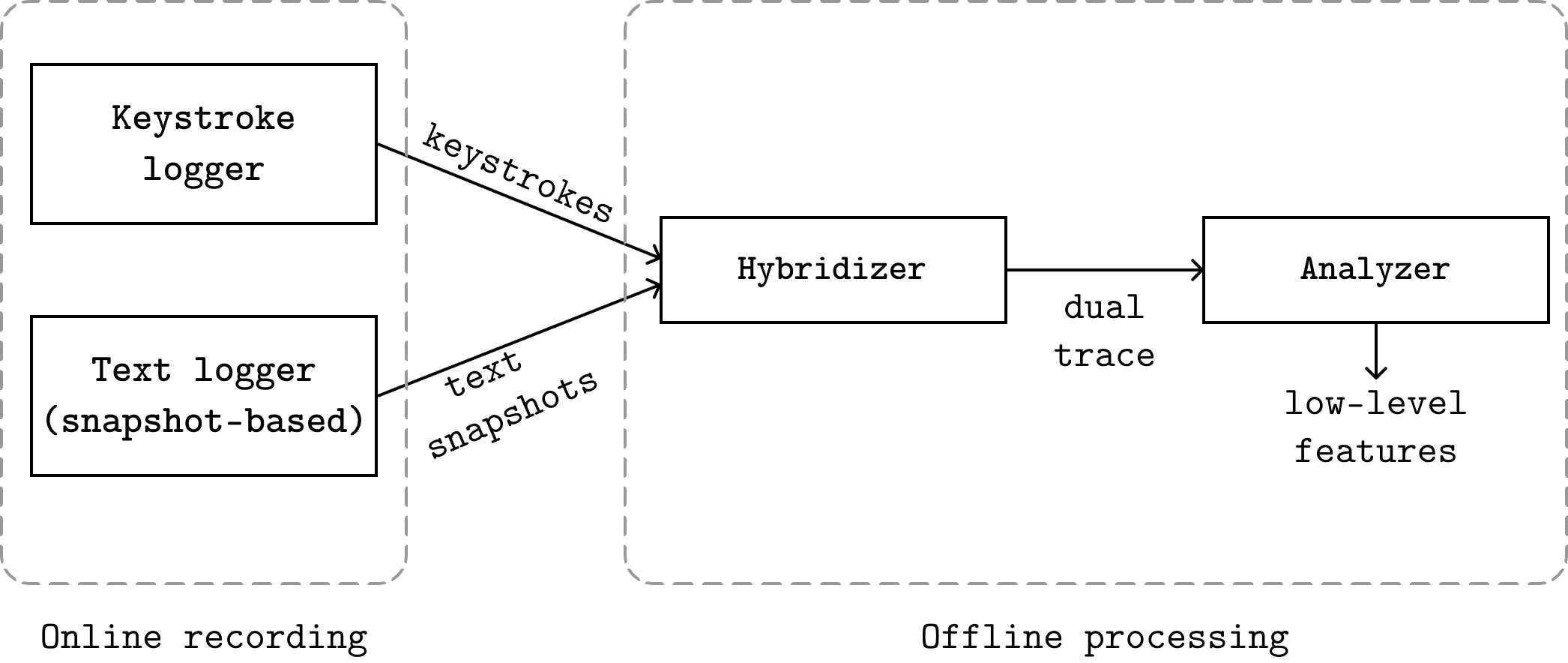}
\caption{Workflow of Hylog, a hybrid input logger.}
\label{fig:workflow}
\end{figure*}

This section presents a hybrid approach that integrates keylogging and text logging to analyze text-production processes in both alphabetic and non-alphabetic scripts. Figure \ref{fig:workflow} outlines the workflow in which a keylogger and a text logger simultaneously monitor the typists' text-production sessions.

Both loggers operate simultaneously and unobtrusively, capturing keystrokes and incremental text snapshots as typists perform their regular text-production tasks using their preferred tools. Keylogging is carried out by Inputlog, while our newly developed ecological text logger, Hylog, records the text as rendered on screen in an ecological fashion. After the session, Hylog's \textit{hybridizer} module post-processes the data from both loggers offline and merges them into a unified dual-track log, enabling cross-validation between keystroke timing and text-state evolution. An \textit{analyzer} module then extracts the relevant metrics. Below, we present the design of Hylog and the main algorithms of the hybridizer module---the core technical contributions of this paper. Finally, we outline the metrics produced by Hylog's analyzer module using the merged dual track.

\subsection{Ecological text logging}

Two main challenges arose in designing Hylog. First, typists frequently alternate between multiple tools and documents. For example, when translating, a typist may work with their preferred text processor, produce their main text in a document, annotate some separate source documents and occasionally enter text into browser windows to search for online information. Therefore, text loggers should capture text production across multiple sources, applications, and environments. We addressed this by designing Hylog as a modular suite of plug-ins, each dedicated to a supported text-entering environment. Our current pilot implementation of Hylog includes a plug-in for Microsoft Word and a plug-in for the Google Chrome browser, both for simplified Chinese.

Second, the typists may routinely edit scattered text excerpts, which introduces performance challenges related to handling large documents, and difficulties in tracking which document excerpt the typist is actively editing. This particularly impacts text-editor plug-ins, like the one for Microsoft Word. To address this, we devised an algorithm to log suitably selected stretches of the target document—labeled \emph{snapshots}—by dynamically tracking the points where it is being manipulated at each logging instant.

Non-ecological input loggers dodge these challenges by restricting the typists' work area to a single internal input window, thus recording only a small fraction of their actual text-production activities. Our logging strategy for Hylog employs high-frequency text sampling and a \textit{dynamic snapshot window} (DSW) algorithm for efficient handling of large documents. Below we describe this approach and its implementation in Microsoft Word and Google Chrome plug-ins.

\subsubsection{Snapshot-based text logging}\label{sec:snapshot-based-text-logging}

Hylog samples text at high frequency and logs \emph{snapshots} that represent only the differences between consecutive versions of the text. This approach increases efficiency by avoiding redundant storage of the entire text at every sampling pass. Formally, given a text, evolving over passes $i$ with $i \in \mathbb{N}$, starting from the initial pass $i_0$, the snapshot at each pass $i$ is defined in \eqref{eq:snapshot} as follows:

\begin{equation}
snapshot(i) =\begin{cases}
\text{initial text} & \text{if } i = i_0 \\
\delta(text(i), text(i-1)) & \text{if } i > i_0
\end{cases}\label{eq:snapshot}
\end{equation}

\noindent where $\delta(new\_text, old\_text)$ is a function that computes the difference between the two texts. To ensure efficiency, this is implemented using a standard diff algorithm, such as the Myers' diff algorithm.

The plug-in queries the writing tool at each pass to retrieve the current text, and keeps the previous text (state) in memory to be able to compute the differences. The output of this plug-in is the trace of the snapshots recorded at every pass, each paired with a corresponding timestamp. Thus, based on this output, it is possible to straightforwardly reconstruct the status of the entire text at each pass, as showed in \eqref{eq:text}:

\begin{equation}
text(i) =
\begin{cases}
snapshot(i_0) & \text{if } i = i_0 \\
\sigma(text(i-1), snapshot(i)) & \text{if } i > i_0
\end{cases}
\label{eq:text}
\end{equation}

\noindent where $\sigma(source\_text, snapshot)$ is a text reconstruction function that takes a source text as input and the difference (i.e., the snapshot) with respect to the target text, and gives the target text as output.

This approach efficiently optimizes storage space for logging data, because it records only the changes between consecutive passes, i.e., versions of the unfolding text. A limitation of this method is that the plug-in must retrieve and compare the entire document at each pass, which can be computationally demanding for long texts. Thus, we restrict this strategy to applications that typically handle short text segments. For instance, Hylog's plug-in for the Google Chrome browser implements this strategy, as the text that typists enter in the browser for online surfing and searching is often written from scratch and generally remain limited in size. For large text editors (text processors), we developed an optimized variant of this strategy, described below.

\subsubsection{The dynamic-snapshot-window algorithm}

To efficiently log large documents, we implemented a \textit{dynamic snapshot window} (DSW) algorithm that, instead of comparing full-document versions, retrieves \emph{snapshots} that represent only the text segment actively modified by the typist.

\begin{table*}[]
\centering
\caption{Behavior of the DSW algorithm across six passes (P1--P6) showing DSW bounds (green/red pipes) and cursor position (blue arrow).}
\label{tab:dsw-moving-example}
\resizebox{0.7\linewidth}{!}{
\begin{tabular}{clllllllllllllllll}
\toprule
\multicolumn{1}{c}{} & \multicolumn{1}{c}{\textbf{0}} & \multicolumn{1}{c}{\textbf{1}} & \multicolumn{1}{c}{\textbf{2}} & \multicolumn{1}{c}{\textbf{3}} & \multicolumn{1}{c}{\textbf{4}} & \multicolumn{1}{c}{\textbf{5}} & \multicolumn{1}{c}{\textbf{6}} & \multicolumn{1}{c}{\textbf{7}} & \multicolumn{1}{c}{\textbf{8}} & \multicolumn{1}{c}{\textbf{9}} & \multicolumn{1}{c}{\textbf{10}} & \multicolumn{1}{c}{\textbf{11}} & \multicolumn{1}{c}{\textbf{12}} & \multicolumn{1}{c}{\textbf{13}} & \multicolumn{1}{c}{\textbf{14}} & \multicolumn{1}{c}{\textbf{15}} & \multicolumn{1}{c}{\textbf{16}} \\ 
\midrule
\vcell{\textbf{P1}} & \vcell{\begin{tabular}[b]{@{}l@{}}\\A\end{tabular}} & \vcell{} & \vcell{\begin{tabular}[b]{@{}l@{}}\\j\end{tabular}} & \vcell{\begin{tabular}[b]{@{}l@{}}\\o\end{tabular}} & \vcell{\begin{tabular}[b]{@{}l@{}}\\i\end{tabular}} & \vcell{\begin{tabular}[b]{@{}l@{}}\\r\end{tabular}} & \vcell{\begin{tabular}[b]{@{}l@{}}\\n\end{tabular}} & \vcell{\begin{tabular}[b]{@{}l@{}}\\e\end{tabular}} & \vcell{\begin{tabular}[b]{@{}l@{}}\\y\end{tabular}} & \vcell{} & \vcell{\begin{tabular}[b]{@{}l@{}}\\o\end{tabular}} & \vcell{\begin{tabular}[b]{@{}l@{}}\textcolor[rgb]{0.122,0.651,0.012}{}\\\textcolor[rgb]{0.122,0.651,0.012}{\textbar{}\textbar{}}\textcolor{red}{\textbar{}\textbar{}}\\ \textcolor[rgb]{0,0,1}{↑}\end{tabular}} & \vcell{\begin{tabular}[b]{@{}l@{}}\\\\\end{tabular}} & \vcell{\begin{tabular}[b]{@{}l@{}}\\\\\end{tabular}} & \vcell{\begin{tabular}[b]{@{}l@{}}\\\\\end{tabular}} & \vcell{\begin{tabular}[b]{@{}l@{}}\\\\\end{tabular}} & \vcell{\begin{tabular}[b]{@{}l@{}}\\\\\end{tabular}} \\[-\rowheight]
\printcellmiddle & \printcelltop & \printcelltop & \printcelltop & \printcelltop & \printcelltop & \printcelltop & \printcelltop & \printcelltop & \printcelltop & \printcelltop & \printcelltop & \printcelltop & \printcelltop & \printcelltop & \printcelltop & \printcelltop & \printcelltop \\ 
\vcell{\textbf{P2}} & \vcell{\begin{tabular}[b]{@{}l@{}}\\A\end{tabular}} & \vcell{\begin{tabular}[b]{@{}l@{}}\\\end{tabular}} & \vcell{\begin{tabular}[b]{@{}l@{}}\\j\end{tabular}} & \vcell{\begin{tabular}[b]{@{}l@{}}\\o\end{tabular}} & \vcell{\begin{tabular}[b]{@{}l@{}}\\i\end{tabular}} & \vcell{\begin{tabular}[b]{@{}l@{}}\\r\end{tabular}} & \vcell{\begin{tabular}[b]{@{}l@{}}\\n\end{tabular}} & \vcell{\begin{tabular}[b]{@{}l@{}}\\e\end{tabular}} & \vcell{\begin{tabular}[b]{@{}l@{}}\\y\end{tabular}} & \vcell{\begin{tabular}[b]{@{}l@{}}\\\end{tabular}} & \vcell{\begin{tabular}[b]{@{}l@{}}\\o\end{tabular}} & \vcell{\begin{tabular}[b]{@{}l@{}}\textcolor[rgb]{0.122,0.651,0.012}{}\\\textcolor[rgb]{0.122,0.651,0.012}{\textbar{}f}\end{tabular}} & \vcell{\begin{tabular}[b]{@{}l@{}}\\\end{tabular}} & \vcell{\begin{tabular}[b]{@{}l@{}}\textcolor[rgb]{0.122,0.651,0.012}{}\\\textcolor[rgb]{0.122,0.651,0.012}{a}\end{tabular}} & \vcell{\begin{tabular}[b]{@{}l@{}}\\\end{tabular}} & \vcell{\begin{tabular}[b]{@{}l@{}}\textcolor[rgb]{0.122,0.651,0.012}{}\\\textcolor[rgb]{0.122,0.651,0.012}{t}\end{tabular}} & \vcell{\begin{tabular}[b]{@{}l@{}}\textcolor[rgb]{0.122,0.651,0.012}{}\\\textcolor[rgb]{0.122,0.651,0.012}{\textbar{}}\textcolor{red}{\textbar{}\textbar{}}\\ \textcolor[rgb]{0,0,1}{↑}\end{tabular}} \\[-\rowheight]
\printcellmiddle & \printcelltop & \printcelltop & \printcelltop & \printcelltop & \printcelltop & \printcelltop & \printcelltop & \printcelltop & \printcelltop & \printcelltop & \printcelltop & \printcelltop & \printcelltop & \printcelltop & \printcelltop & \printcelltop & \printcelltop \\ 
\vcell{\textbf{P3}} & \vcell{\begin{tabular}[b]{@{}l@{}}\\A\end{tabular}} & \vcell{\begin{tabular}[b]{@{}l@{}}\\\end{tabular}} & \vcell{\begin{tabular}[b]{@{}l@{}}\\j\end{tabular}} & \vcell{\begin{tabular}[b]{@{}l@{}}\\o\end{tabular}} & \vcell{\begin{tabular}[b]{@{}l@{}}\\i\end{tabular}} & \vcell{\begin{tabular}[b]{@{}l@{}}\textcolor[rgb]{0.122,0.651,0.012}{}\\\textcolor{red}{\textbar{}\textbar{}}\textcolor[rgb]{0.122,0.651,0.012}{\textbar{}r} \\ \textcolor[rgb]{0,0,1}{~↑}\end{tabular}} & \vcell{\begin{tabular}[b]{@{}l@{}}\textcolor[rgb]{0.122,0.651,0.012}{}\\\textcolor[rgb]{0.122,0.651,0.012}{n}\end{tabular}} & \vcell{\begin{tabular}[b]{@{}l@{}}\textcolor[rgb]{0.122,0.651,0.012}{}\\\textcolor[rgb]{0.122,0.651,0.012}{e}\end{tabular}} & \vcell{\begin{tabular}[b]{@{}l@{}}\textcolor[rgb]{0.122,0.651,0.012}{}\\\textcolor[rgb]{0.122,0.651,0.012}{y}\end{tabular}} & \vcell{\begin{tabular}[b]{@{}l@{}}\\\end{tabular}} & \vcell{\begin{tabular}[b]{@{}l@{}}\textcolor[rgb]{0.122,0.651,0.012}{}\\\textcolor[rgb]{0.122,0.651,0.012}{o}\end{tabular}} & \vcell{\begin{tabular}[b]{@{}l@{}}\textcolor[rgb]{0.122,0.651,0.012}{}\\\textcolor[rgb]{0.122,0.651,0.012}{f}\end{tabular}} & \vcell{\begin{tabular}[b]{@{}l@{}}\\\end{tabular}} & \vcell{\begin{tabular}[b]{@{}l@{}}\textcolor[rgb]{0.122,0.651,0.012}{}\\\textcolor[rgb]{0.122,0.651,0.012}{a}\end{tabular}} & \vcell{\begin{tabular}[b]{@{}l@{}}\\\end{tabular}} & \vcell{\begin{tabular}[b]{@{}l@{}}\textcolor[rgb]{0.122,0.651,0.012}{}\\\textcolor[rgb]{0.122,0.651,0.012}{t}\end{tabular}} & \vcell{\begin{tabular}[b]{@{}l@{}}\textcolor[rgb]{0.122,0.651,0.012}{}\\\textcolor[rgb]{0.122,0.651,0.012}{\textbar{}}\end{tabular}} \\[-\rowheight]
\printcellmiddle & \printcelltop & \printcelltop & \printcelltop & \printcelltop & \printcelltop & \printcelltop & \printcelltop & \printcelltop & \printcelltop & \printcelltop & \printcelltop & \printcelltop & \printcelltop & \printcelltop & \printcelltop & \printcelltop & \printcelltop \\ 
\vcell{\textbf{P4}} & \vcell{\begin{tabular}[b]{@{}l@{}}\\A\end{tabular}} & \vcell{\begin{tabular}[b]{@{}l@{}}\\\end{tabular}} & \vcell{\begin{tabular}[b]{@{}l@{}}\\j\end{tabular}} & \vcell{\begin{tabular}[b]{@{}l@{}}\\o\end{tabular}} & \vcell{\begin{tabular}[b]{@{}l@{}}\textcolor[rgb]{0.122,0.651,0.012}{}\\\textcolor[rgb]{0.122,0.651,0.012}{\textbar{}u}\end{tabular}} & \vcell{\begin{tabular}[b]{@{}l@{}}\textcolor[rgb]{0.122,0.651,0.012}{}\\\textcolor[rgb]{0.122,0.651,0.012}{\textbar{}}\textcolor{red}{\textbar{}\textbar{}}r \\ \textcolor[rgb]{0,0,1}{~↑}\end{tabular}} & \vcell{\begin{tabular}[b]{@{}l@{}}\\n\end{tabular}} & \vcell{\begin{tabular}[b]{@{}l@{}}\\e\end{tabular}} & \vcell{\begin{tabular}[b]{@{}l@{}}\\y\end{tabular}} & \vcell{\begin{tabular}[b]{@{}l@{}}\\\end{tabular}} & \vcell{\begin{tabular}[b]{@{}l@{}}\\o\end{tabular}} & \vcell{\begin{tabular}[b]{@{}l@{}}\\f\end{tabular}} & \vcell{\begin{tabular}[b]{@{}l@{}}\\\end{tabular}} & \vcell{\begin{tabular}[b]{@{}l@{}}\\a\end{tabular}} & \vcell{\begin{tabular}[b]{@{}l@{}}\\\end{tabular}} & \vcell{\begin{tabular}[b]{@{}l@{}}\\t\end{tabular}} & \vcell{\begin{tabular}[b]{@{}l@{}}\\\end{tabular}} \\[-\rowheight]
\printcellmiddle & \printcelltop & \printcelltop & \printcelltop & \printcelltop & \printcelltop & \printcelltop & \printcelltop & \printcelltop & \printcelltop & \printcelltop & \printcelltop & \printcelltop & \printcelltop & \printcelltop & \printcelltop & \printcelltop & \printcelltop \\ 
\vcell{\textbf{P5}} & \vcell{\begin{tabular}[b]{@{}l@{}}\\A\end{tabular}} & \vcell{\begin{tabular}[b]{@{}l@{}}\\\end{tabular}} & \vcell{\begin{tabular}[b]{@{}l@{}}\textcolor[rgb]{0.122,0.651,0.012}{}\\\textcolor{red}{\textbar{}}\textcolor[rgb]{0.122,0.651,0.012}{\textbar{}j}\end{tabular}} & \vcell{\begin{tabular}[b]{@{}l@{}}\textcolor[rgb]{0.122,0.651,0.012}{}\\\textcolor[rgb]{0.122,0.651,0.012}{o}\end{tabular}} & \vcell{\begin{tabular}[b]{@{}l@{}}\textcolor[rgb]{0.122,0.651,0.012}{}\\\textcolor[rgb]{0.122,0.651,0.012}{u}\end{tabular}} & \vcell{\begin{tabular}[b]{@{}l@{}}\textcolor[rgb]{0.122,0.651,0.012}{}\\\textcolor[rgb]{0.122,0.651,0.012}{~r}\end{tabular}} & \vcell{\begin{tabular}[b]{@{}l@{}}\textcolor[rgb]{0.122,0.651,0.012}{}\\\textcolor[rgb]{0.122,0.651,0.012}{n}\end{tabular}} & \vcell{\begin{tabular}[b]{@{}l@{}}\textcolor[rgb]{0.122,0.651,0.012}{}\\\textcolor[rgb]{0.122,0.651,0.012}{e}\end{tabular}} & \vcell{\begin{tabular}[b]{@{}l@{}}\textcolor[rgb]{0.122,0.651,0.012}{}\\\textcolor[rgb]{0.122,0.651,0.012}{y}\end{tabular}} & \vcell{\begin{tabular}[b]{@{}l@{}}\textcolor[rgb]{0.122,0.651,0.012}{}\\\textcolor[rgb]{0.122,0.651,0.012}{\textbar{}}\textcolor{red}{\textbar{}} \\ \textcolor[rgb]{0,0,1}{↑}\end{tabular}} & \vcell{\begin{tabular}[b]{@{}l@{}}\\o\end{tabular}} & \vcell{\begin{tabular}[b]{@{}l@{}}\\f\end{tabular}} & \vcell{\begin{tabular}[b]{@{}l@{}}\\\end{tabular}} & \vcell{\begin{tabular}[b]{@{}l@{}}\\a\end{tabular}} & \vcell{\begin{tabular}[b]{@{}l@{}}\\\end{tabular}} & \vcell{\begin{tabular}[b]{@{}l@{}}\\t\end{tabular}} & \vcell{\begin{tabular}[b]{@{}l@{}}\\\end{tabular}} \\[-\rowheight]
\printcellmiddle & \printcelltop & \printcelltop & \printcelltop & \printcelltop & \printcelltop & \printcelltop & \printcelltop & \printcelltop & \printcelltop & \printcelltop & \printcelltop & \printcelltop & \printcelltop & \printcelltop & \printcelltop & \printcelltop & \printcelltop \\ 
\vcell{\textbf{P6}} & \vcell{\begin{tabular}[b]{@{}l@{}}\\A\end{tabular}} & \vcell{\begin{tabular}[b]{@{}l@{}}\\\\\end{tabular}} & \vcell{\begin{tabular}[b]{@{}l@{}}\textcolor[rgb]{0.122,0.651,0.012}{}\\\textcolor[rgb]{0.122,0.651,0.012}{\textbar{}t}\end{tabular}} & \vcell{\begin{tabular}[b]{@{}l@{}}\textcolor[rgb]{0.122,0.651,0.012}{}\\\textcolor[rgb]{0.122,0.651,0.012}{r}\end{tabular}} & \vcell{\begin{tabular}[b]{@{}l@{}}\textcolor[rgb]{0.122,0.651,0.012}{}\\\textcolor[rgb]{0.122,0.651,0.012}{a}\end{tabular}} & \vcell{\begin{tabular}[b]{@{}l@{}}\textcolor[rgb]{0.122,0.651,0.012}{}\\\textcolor[rgb]{0.122,0.651,0.012}{v}\end{tabular}} & \vcell{\begin{tabular}[b]{@{}l@{}}\textcolor[rgb]{0.122,0.651,0.012}{}\\\textcolor[rgb]{0.122,0.651,0.012}{e}\end{tabular}} & \vcell{\begin{tabular}[b]{@{}l@{}}\textcolor[rgb]{0.122,0.651,0.012}{}\\\textcolor[rgb]{0.122,0.651,0.012}{l}\end{tabular}} & \vcell{\begin{tabular}[b]{@{}l@{}}\textcolor[rgb]{0.902,0,0.055}{}\\\textcolor{red}{\textbar{}\textbar{}}\\ \textcolor[rgb]{0,0,1}{↑}\end{tabular}} & \vcell{\begin{tabular}[b]{@{}l@{}}\textcolor[rgb]{0.122,0.651,0.012}{}\\\textcolor[rgb]{0.122,0.651,0.012}{\textbar{}\textcolor{black}{o}}\end{tabular}} & \vcell{\begin{tabular}[b]{@{}l@{}}\\f\end{tabular}} & \vcell{\begin{tabular}[b]{@{}l@{}}\\\\\end{tabular}} & \vcell{\begin{tabular}[b]{@{}l@{}}\\a\end{tabular}} & \vcell{\begin{tabular}[b]{@{}l@{}}\\\\\end{tabular}} & \vcell{\begin{tabular}[b]{@{}l@{}}\\t\end{tabular}} & \vcell{\begin{tabular}[b]{@{}l@{}}\\\\\end{tabular}} & \vcell{\begin{tabular}[b]{@{}l@{}}\\\\\end{tabular}} \\[-\rowheight]
\printcellmiddle & \printcelltop & \printcelltop & \printcelltop & \printcelltop & \printcelltop & \printcelltop & \printcelltop & \printcelltop & \printcelltop & \printcelltop & \printcelltop & \printcelltop & \printcelltop & \printcelltop & \printcelltop & \printcelltop & \printcelltop \\
\bottomrule
\end{tabular}
}
\end{table*}

We illustrate Hylog's DSW algorithm with an example showing how it behaves as a typist produces a short text in English. In this version of the algorithm, each recorded snapshot corresponds to a portion of the document. This is the \emph{dynamic snapshot window} (DSW). During a logging session, the DSW moves through the document like a sliding window, but it also shrinks and expands, such that the snapshots capture the relevant changes made by the typist on the text.

Table \ref{tab:dsw-moving-example} shows a text as it unfolds through six consecutive passes, $P1$--$P6$, with both additions and corrections. Each row represents a pass and the corresponding incremental snapshot recorded by Hylog. In the example, we assume that there was some text-production activity before this session, so that when opening a document at the beginning some text is already present. A copy of the document is created right away, implementing pass $P0$, which captures the initial text, i.e., $snapshot(P0)$ = `A joirney o'. We use the following conventions: green highlighting shows the text in focus within the DSW (i.e., the text that will be captured with a snapshot); a blue arrow indicates the cursor position; pipe symbols mark the DSW boundaries before (green pipes) and after (red pipes) taking the snapshot: 

\begin{enumerate}

\item In $P1$, the text `A joirney o' appears in the range [0,11) of the letters and the cursor is at the end of the document, i.e., at position 11. The DSW starts and ends at position 11, indicating that there is nothing to log, as no additional text has been entered yet. With this setting of the DSW, the snapshot taken at $P1$ is empty and the DSW remains the same after taking the snapshot.

\item Between $P1$ and $P2$, the text `f a t' is typed. Hylog's algorithm detects that the cursor position increased with respect to the right bound of the DSW after $P1$, and expands the DSW rightward, covering the text portion [11, 16), which matches the added text. The snapshot taken at $P2$ queries the text processor for the data within the DSW, that is, the portion of text [11, 16), and records the text `f a t' in the log. After taking the snapshot, the bounds of the DSW are updated to match the current position of the cursor (16), preparing to repeat the logging process at the next pass.

\item Between $P2$ and $P3$ the typist realizes she mistyped `jo\textbf{i}rney' (instead of `jo\textbf{u}rney'). To correct the typo, she sets the cursor at position 5, just before the `r'. Mirroring the behavior described above, at $P3$ the plug-in detects that the cursor position decreased at the left bound of the DSW after $P2$ and expands the DSW leftward, so that the snapshot taken at $P3$ captures all the letters in the range [5, 16). In this case, the text has not been modified, but Hylog's algorithm conservatively captures all text subtended by the cursor movement. After taking the snapshot, the bounds of the DSW are updated to match the new cursor position (5).

\item Between $P3$ and $P4$, \texttt{BACKSPACE} is pressed once, causing the letter `i' to disappear, and then the letter `u' is entered. Hylog's algorithm intercepts the number of backspaces, and moves the left bound of the DSW of a corresponding amount, thus putting position 4 under observation. As a consequence, the snapshot taken at $P4$ retrieves the text [4, 5) and records only `u' in the log. Knowing the new status at the interval [4, 5)—letter `u'—is enough to reconstruct the evolution of the text, based on the logged data. Then, as above, the bounds of the DSW are updated to match the position of the cursor (5).

\item Between $P4$ and $P5$ the typist double clicks and selects the word `journey'. The selection event brings the cursor at the end of the highlighted word, i.e., at position 9, and indicates that the selection starts at position 2, leading the algorithm to adjust the DSW bounds such that they coincide with the selection, focusing on the range [2, 9). Thus, the snapshot taken at $P5$ contains the selected text. Since the text remains selected after the snapshot, the DSW remains on [2, 9).

\item Between $P5$ and $P6$, the typist replaces the selected text by entering `travel'. The algorithm sets the DSW to cover the larger interval between either the selected text or the newly typed text: in this case, the selected text is at the interval [2,9), and the typed text is at the interval [2,8); thus, the DSW remains [2,9). In this way, the snapshot taken at $P6$ necessarily captures all replaced letters. After the snapshot, the DSW is updated to [8,8) according to the cursor position.

\end{enumerate} 

In addition to what the table shows, Hylog's plug-in also detects that the document's overall length decreased by one letter and logs this offset information ($offset=-1$), which is necessary for reconstructing document evolution. When inserted text exceeds the replaced selection, the offset computation also accounts for \textit{delete} key presses (\texttt{CANC} and \texttt{BACKSPACE} keys), which the plug-in tracks explicitly.

Algorithm \ref{alg:snapshot_algorithm} formalizes how Hylog controls the DSW and builds snapshots at each timestamp. As input, it takes (a) the current DSW bounds; (b) the cursor position; (c) the start index of any current selection; (d) the count of \texttt{BACKSPACE} and \texttt{CANC} key presses since the last snapshot, and (e) the length of the document. Hylog's plug-in maintains the DSW bounds internally and retrieves the cursor position, selection start, \texttt{BACKSPACE} and \texttt{CANC} key press counts, and document length from the text processor, monitoring key events via the operating system.

The algorithm outputs a snapshot containing the recorded text, its timestamp, and the offset and DSW data needed for precise text reconstruction. These data define the relationship between each snapshot and the evolving document. The algorithm also returns the updated DSW to serve as input for the next snapshot.

\begin{algorithm}[]

\caption{Hylog's algorithm regulates the DSW bounds and takes a snapshot of the text accordingly.}
	\label{alg:snapshot_algorithm}

\SetKwFunction{GetText}{GetText}
\SetKwFunction{GetCurrentTimestamp}{GetCurrentTimestamp}
\SetKwFunction{ComputeOffset}{ComputeOffset}

\Input{
 \textit{dsw = $\langle left, right \rangle$}: the most recent DSW\newline
	\textit{cPos}: the current position of the cursor\newline
	\textit{selectionStart}: the starting position of the text selection\newline
 \textit{backCounter}: the number of times the \texttt{BACKSPACE} key has been pressed since the last snapshot\newline
 \textit{cancCounter}: the number of times the \texttt{CANC} key has been pressed since the last snapshot\newline
 \textit{docLength}: the total number of symbols (either Latin  Letters or Chinese characters) in the document
}

\BlankLine

\Output{
 \textit{snapshot = $\langle text, timestamp, dsw, offset \rangle$}: the latest snapshot of the text \newline
 \textit{dsw' = $\langle left, right\rangle$}: the DSW updated after the snapshot is taken
 }

\BlankLine

\textit{dsw.left} $\leftarrow \min(dsw.left, selectionStart)$ \label{alg:snapshot_algorithm:updating_min:begin}
 
\textit{dsw.left} $\leftarrow \min(dsw.left - backCounter, cPos)$\label{alg:snapshot_algorithm:updating_min:end}

\textit{dsw.left} $\leftarrow \max(0, dsw.left)$ \;

\BlankLine

\textit{dsw.right} $\leftarrow \max(dsw.right, cPos)$ \label{alg:snapshot_algorithm:updating_max}

\BlankLine

\BlankLine

\If{\textit{dsw.left} $=$ \textit{dsw.right} $\land$ \textit{cancCounter} $=$ 0 $\land$ \textit{backCounter} $=$ 0}{
\label{alg:snapshot_algorithm:taking_snapshot:begin}
 \textit{snapshot} $\leftarrow \langle~\rangle$\;\label{alg:snapshot_algorithm:taking_snapshot:nosnapshot}
} \Else{
 \textit{text} $\leftarrow$ \GetText{\textit{dsw.left, dsw.right}}\;
\textit{offset} $\leftarrow$ \ComputeOffset{\textit{cancCounter}, \textit{docLength}}\label{alg:snapshot_algorithm:computing_offset}\;
 \textit{timestamp} $\leftarrow$ \GetCurrentTimestamp()\;
 \textit{snapshot} $\leftarrow$ \textit{$\langle text, offset, timestamp, dsw \rangle$}\;
}\label{alg:snapshot_algorithm:taking_snapshot:end}

\BlankLine

\textit{dsw'.left} $\leftarrow \min(cPos, selectionStart)$\;\label{alg:snapshot_algorithm:updating_dsw:begin}
\textit{dsw'.right} $\leftarrow$ cPos\;\label{alg:snapshot_algorithm:updating_dsw:end}

\BlankLine

\Return{snapshot, dsw'}\label{alg:snapshot_algorithm:return}

\BlankLine

\Fn{\ComputeOffset{cancCounter, docLength}}{
\Return{a number representing the left-shift or right-shift offset of the text}
}

\Fn{\GetText{left, right}}{
\Return{The text contained in the document, starting from the left position (included) and going through the right position (excluded)}
}

\Fn{\GetCurrentTimestamp{}}{
\Return{The current timestamp with millisecond resolution}
}
\end{algorithm}

Hylog's DSW algorithm proceeds through the following steps:

\begin{enumerate}
 
 \item \textbf{Tuning the DSW} \textit{(lines \ref{alg:snapshot_algorithm:updating_min:begin}-\ref{alg:snapshot_algorithm:updating_max})}: first, both DSW bounds (leftward and rightward) are updated, based on the current position of the cursor, the number of backspace operations that occurred since the last snapshot, and the beginning of the active selection, if any. The left bound is never set less than zero. Tracking backspace operations ensures that no deletions are missed during reconstruction. For instance, imagine that the typist has written `h\textbf{a}llo' up to the last taken snapshot. Then, before the next snapshot occurs, \texttt{BACKSPACE} is pressed four times and the word is rewritten as \textit{hello}. In this case, at the time of the second snapshot, the cursor appears at the same position, but \textit{dsw.left} is still moved backwards by four positions at line~\ref{alg:snapshot_algorithm:updating_min:end}, allowing for capturing the replacement of the letter `a' with the `e'.
 
 \item \textbf{Taking the text snapshot} \textit{(lines \ref{alg:snapshot_algorithm:taking_snapshot:begin}-\ref{alg:snapshot_algorithm:taking_snapshot:end})}: if the DSW is empty and neither left nor right cancellations occurred, then no snapshot needs to be taken (line~\ref{alg:snapshot_algorithm:taking_snapshot:nosnapshot}).
 Otherwise, the text between the DSW bounds is retrieved and the offset computed for the reconstruction. Both are saved in the snapshot, along with the current timestamp and DSW data. The offset indicates that the text before the snapshot must be shifted when reconstructing the document. It is calculated by considering deletions happening to the right of the cursor (\texttt{CANC} key presses) and the current length of the document. Cancellations require left shifting, because of the symbols being removed, while added characters require right shifting.
 
 \item \textbf{Updating the DSW after taking the snapshot} \textit{(lines \ref{alg:snapshot_algorithm:updating_dsw:begin}-\ref{alg:snapshot_algorithm:updating_dsw:end})}: Hylog's DSW is prepared for the next snapshot. In most cases, this leads to set the left and right bounds of the DSW at the cursor position, but the left bound can differ to fit a text selection. 

 \item \textbf{Registering the snapshot in the log}: at the end \textit{(line \ref{alg:snapshot_algorithm:return})}, the snapshot is returned to be registered into the log, along with the updated DSW, ready to be used as input for the next iteration of the algorithm.
 
\end{enumerate}

Snapshots captured through Hylog's DSW algorithm make it possible to reconstruct the \textit{full} state of the text at any given time.

\begin{rev}
The DSW algorithm addresses concurrent editings as follows:

\begin{itemize}
\item  it relies on the API of the word processor at hand that serializes the graphical events. This guarantees that selection events cannot overlap, even if the user moves the cursor across the document with high frequency;
\item it uses a mutex lock to prevent overlapping iterations of the algorithm, if the time-interval timers fire before the conclusion of the ongoing iteration. This ensures mutual exclusion and prevents concurrent updates;
\item it checks the correct ordering of graphical and timer events. In case of incorrect orderings, it deletes the snapshots that were inconistently recorded, 
rolls back the DSW to the state before the inconsistency, and triggers a new iteration of the algorithm, in order to log the correct snapshot.
\end{itemize}
\end{rev}

\subsubsection{Text logging plug-in implementations}

\paragraph{Wordtextlogger} Hylog's DSW plug-in was implemented as a Microsoft Word add-in. We used the Visual Studio Tools for Office framework, or VSTO \cite{vsto}, and accessed \texttt{C}, \texttt{C++} and \texttt{C\#} Microsoft APIs for Word, Office, and the operating system. When active, Hylog begins recording whenever a document file is opened or created, and saves the output in a JSON format, at regular intervals.

A multi-threaded architecture prevents interface lag and triggers periodic snapshots based on timers or user events such as cursor movement. In our implementation, Hylog's Word plug-in samples text at approximately 500 ms intervals in a best-effort fashion. An error-correction mechanism ensures both chronological integrity, by filtering the snapshots for which the event order is uncertain, and structural integrity by triggering additional snapshots in case of user actions that may impact on the structure of the text, such as cursor moves and text selections. 

\paragraph{Chrometextlogger} Hylog's plug-in for Google Chrome is a JavaScript Extension. Active during web browsing, it scans accessed pages to detect input fields and capture snapshots of any text typed in. This plug-in applies the baseline snapshot-based algorithm outlined above. By contrast, it relies on input-field event listeners and records updates only when the text changes as a direct result of user input. 

\begin{rev}
\subsubsection{\begin{rev} Performance of the DSW algorithm \end{rev}}\label{sec:dsw:performance}
To evaluate the performance of the DSW algorithm we have conducted a controlled experiment in which we compared our implmentation of the Wordtextlogger based on the DSW algorithm with 
the baseline snapshot-based text logging (§~\ref{sec:snapshot-based-text-logging}). We aimed at assessing both that the DSW algorithm outperforms the baseline approach, and that it offers the required level of unobtrusiveness when users shall deal with large documents.

For the experiment we implemented a version of Hylog equipped with the snapshot-based logging:
every 500 milliseconds or when selection event occurs, it retrieves all text in the document, calculates the differences with respect to the text captured at the previous iteration, and saves the differences into a log file. 

We generated a set of 5 documents of increasing lengths, by truncating an existing book at 140, 160, 180 and 200 pages, and also considering the entire book of 350 pages.\footnote{\begin{rev}We used non-copyrighted version of the novel ``I promessi sposi'' by Alessandro Manzoni, formatted with Word default font Aptos 12pt (body).\end{rev}}

For each document, we executed two writing sessions of 10 minutes in which we monitored the process with Hylog equipped with either the DSW algorithm or the baseline snapshot-based algorithm, respectively. In each writing session, we repeated the following experimental protocol:

\begin{enumerate}
    \item Open the document;
    \item Scroll to the bottom of the document using the lateral bar;
    \item Select the last three lines with the mouse and hit backspace on the keyboard;
    \item Type a text at the end the document: specifically, we typed the first three paragraphs of another book (``Harry Potter and the Philosopher's Stone'');
    \item Scroll to the beginning of the document using the lateral bar;
    \item Type a few lines of free text about today's weather;
    \item Scroll to the middle page of the document using the lateral bar;
    \item Type a few lines of free text about yesterday's weather;
    \item Save the document;
    \item Close the document.
\end{enumerate}

At the end of session, we annotated our qualitative evaluation about the reactivity of the user interface at the loading of the document, during the production of text, when scrolling the document, when moving the cursor across the document, and when saving and closing the document.
We paused after each session to allow the CPU and memory to recover.

Table~\ref{tab:dsw:benchmark} summarizes the results.
The cells of the table are colored as follows: green indicates that we did not perceive any impact of Hylog or its baseline version on the reactivity of the user interface, yellow indicates low impact, orange indicates high impact and red indicates that the user interface became unresponsive.

While the session with the 140-page document did not reveal differences between Hylog and the baseline, as both approaches performed effectively, the sessions with larger documents confirmed the benefits of the DSW algorithm. 
The performance of the baseline incrementally worsens: with documents of 160 pages, the reactivity of the user interface is affected by non-blocking lags while typing text or scrolling the document; with 180 pages, even if it is still possible to use the word processor, the lags heavily compromise the user experience; eventually, Word completely stops responding after loading a document with 200 pages or more, and we had to terminate the application with the Windows Task Manager.  

On the other hand, with the DSW algorithm, all sessions were completed smoothly without obtrusive impact from Hylog. 
We observed a tiny delay only with the 350-page document, when the user brings the cursor very far away from its previous position, e.g., moving from page number 1 directly to page 350. Indeed this is the worst case for the DSW algorithm, in which the window boundaries span all the text subtended by the cursor jump, making Hylog capture the entire document.
While this is needed to avoid losses, moving the cursor through the entire document is a rare event, and caused an observable tiny delay only for the maximum-size document considered in our experiment.

\begin{revv}
We acknowledge that the benchmark we presented validates the proposed DSW algorithm under controlled conditions and it does not stress the plugin in a real-world scenario. However, we consider this result as satisfactory for this prototype version of Hylog and we leave further investigation for future work.
\end{revv}

\begin{table*}
\begin{rev}
\centering
\setlength{\extrarowheight}{0pt}
\addtolength{\extrarowheight}{\aboverulesep}
\addtolength{\extrarowheight}{\belowrulesep}
\setlength{\aboverulesep}{0pt}
\setlength{\belowrulesep}{0pt}
\caption{Qualitative performance comparison between snapshot-based diff logger and Wordtextlogger implementing the DSW algorithm.}
\begin{tabular}{lcccccccccc} 
\toprule
\multirow{2}{*}{} & \multicolumn{2}{c}{\textbf{140 pages}} & \multicolumn{2}{c}{\textbf{160~\textbf{pages}}} & \multicolumn{2}{c}{\textbf{180~\textbf{pages}}} & \multicolumn{2}{c}{\textbf{200~\textbf{pages}}} & \multicolumn{2}{c}{\textbf{350~\textbf{pages}}} \\ 
\cmidrule{2-11}
 & \textbf{Base} & \textbf{DSW} & \textbf{\textbf{Base}} & \textbf{\textbf{DSW}} & \textbf{\textbf{Base}} & \textbf{DSW} & \textbf{\textbf{\textbf{\textbf{Base}}}} & \textbf{\textbf{DSW}} & \textbf{\textbf{Base}} & \textbf{\textbf{DSW}} \\
\begin{tabular}[c]{@{}l@{}}\textbf{UI}\\\textbf{reactivity}\\\textbf{when}\\\textbf{document}\\\textbf{loads}\end{tabular} & \cellcolor[rgb]{0.478,1,0.243}Regular & \cellcolor[rgb]{0.478,1,0.243}Regular & {\cellcolor[rgb]{1,0.941,0.275}}\begin{tabular}[c]{@{}>{\cellcolor[rgb]{1,0.941,0.275}}c@{}}Quite\\regular\end{tabular} & {\cellcolor[rgb]{0.478,1,0.243}}Regular & {\cellcolor[rgb]{0.984,0.62,0.035}}\textcolor{white}{Slow} & {\cellcolor[rgb]{0.478,1,0.243}}Regular & {\cellcolor[rgb]{0.992,0.506,0.482}}\begin{tabular}[c]{@{}>{\cellcolor[rgb]{0.992,0.506,0.482}}c@{}}\textcolor{white}{Cursor not}\\\textcolor{white}{blinking}\\\textcolor{white}{and UI}\\~\textcolor{white}{freezed}\end{tabular} & {\cellcolor[rgb]{0.478,1,0.243}}Regular & {\cellcolor[rgb]{0.992,0.506,0.482}}\begin{tabular}[c]{@{}>{\cellcolor[rgb]{0.992,0.506,0.482}}c@{}}\textcolor{white}{Cursor not}\\\textcolor{white}{blinking}\\\textcolor{white}{and~UI~}\\\textcolor{white}{freezed}\end{tabular} & {\cellcolor[rgb]{0.478,1,0.243}}Regular \\
\begin{tabular}[c]{@{}l@{}}\textbf{Text}\\\textbf{production}\end{tabular} & \cellcolor[rgb]{0.478,1,0.243}Fluent & \cellcolor[rgb]{0.478,1,0.243}Fluent & {\cellcolor[rgb]{1,0.941,0.275}}\begin{tabular}[c]{@{}>{\cellcolor[rgb]{1,0.941,0.275}}c@{}}Lags:~\\delay about \\below \\a second\end{tabular} & {\cellcolor[rgb]{0.478,1,0.243}}Fluent & {\cellcolor[rgb]{0.984,0.62,0.035}}\begin{tabular}[c]{@{}>{\cellcolor[rgb]{0.984,0.62,0.035}}c@{}}\textcolor{white}{Heavy lags:}~\\\textcolor{white}{delay can }\\\textcolor{white}{be above}\\\textcolor{white}{one second}\end{tabular} & {\cellcolor[rgb]{0.478,1,0.243}}Fluent & {\cellcolor[rgb]{0.992,0.506,0.482}}\textcolor{white}{Not possible} & {\cellcolor[rgb]{0.478,1,0.243}}Fluent & {\cellcolor[rgb]{0.992,0.506,0.482}}\textcolor{white}{Not possible} & {\cellcolor[rgb]{0.478,1,0.243}}Fluent \\
\begin{tabular}[c]{@{}l@{}}\textbf{Scroll with}\\\textbf{lateral bar}\end{tabular} & \cellcolor[rgb]{0.478,1,0.243}Smooth & \cellcolor[rgb]{0.478,1,0.243}Smooth & {\cellcolor[rgb]{1,0.941,0.275}}\begin{tabular}[c]{@{}>{\cellcolor[rgb]{1,0.941,0.275}}c@{}}Enough \\smooth\end{tabular} & {\cellcolor[rgb]{0.478,1,0.243}}Smooth & {\cellcolor[rgb]{0.984,0.62,0.035}}\textcolor{white}{Heavy lags} & {\cellcolor[rgb]{0.478,1,0.243}}Smooth & {\cellcolor[rgb]{0.992,0.506,0.482}}\textcolor{white}{Not possible} & {\cellcolor[rgb]{0.478,1,0.243}}Smooth & {\cellcolor[rgb]{0.992,0.506,0.482}}\textcolor{white}{Not possible} & {\cellcolor[rgb]{0.478,1,0.243}}Smooth \\
\begin{tabular}[c]{@{}l@{}}\textbf{Cursor}\\\textbf{movements}\\\textbf{and jumps}\end{tabular} & \cellcolor[rgb]{0.478,1,0.243}Regular & \cellcolor[rgb]{0.478,1,0.243}Regular & {\cellcolor[rgb]{0.478,1,0.243}}Regular & {\cellcolor[rgb]{0.478,1,0.243}}Regular & {\cellcolor[rgb]{0.478,1,0.243}}Regular & {\cellcolor[rgb]{0.478,1,0.243}}Regular & {\cellcolor[rgb]{0.992,0.506,0.482}}\textcolor{white}{Not possible} & {\cellcolor[rgb]{0.478,1,0.243}}Regular & {\cellcolor[rgb]{0.992,0.506,0.482}}\textcolor{white}{Not possible} & {\cellcolor[rgb]{1,0.941,0.275}}\begin{tabular}[c]{@{}>{\cellcolor[rgb]{1,0.941,0.275}}c@{}}Tiny delay\\~if~cursor is \\brought~\\very far\end{tabular} \\
\begin{tabular}[c]{@{}l@{}}\textbf{Saving and}\\\textbf{closing the}\\\textbf{document}\end{tabular} & \cellcolor[rgb]{0.478,1,0.243}Graceful & \cellcolor[rgb]{0.478,1,0.243}Graceful & {\cellcolor[rgb]{0.478,1,0.243}}Graceful & {\cellcolor[rgb]{0.478,1,0.243}}Graceful & {\cellcolor[rgb]{0.478,1,0.243}}Graceful & {\cellcolor[rgb]{0.478,1,0.243}}Graceful & {\cellcolor[rgb]{0.992,0.506,0.482}}\begin{tabular}[c]{@{}>{\cellcolor[rgb]{0.992,0.506,0.482}}c@{}}\textcolor{white}{Application}\\\textcolor{white}{must be}\\\textcolor{white}{terminated}\end{tabular} & {\cellcolor[rgb]{0.478,1,0.243}}Graceful & {\cellcolor[rgb]{0.992,0.506,0.482}}\begin{tabular}[c]{@{}>{\cellcolor[rgb]{0.992,0.506,0.482}}c@{}}\textcolor{white}{Application}\\\textcolor{white}{must be}\\\textcolor{white}{terminated}\end{tabular} & {\cellcolor[rgb]{0.478,1,0.243}}Graceful \\
\bottomrule
\end{tabular}\label{tab:dsw:benchmark}
\end{rev}
\end{table*}

\end{rev}

\subsection{Hybrid keylogging}
Hylog's dual trace enables quantitative analysis of text-production phenomena in non-alphabetic scripts: linking the fine-grained input dynamics of typist actions with the rendered IME output improves behavioral and cognitive interpretation.

\begin{figure*}[h]
\centering
\includegraphics[width=0.7\textwidth]{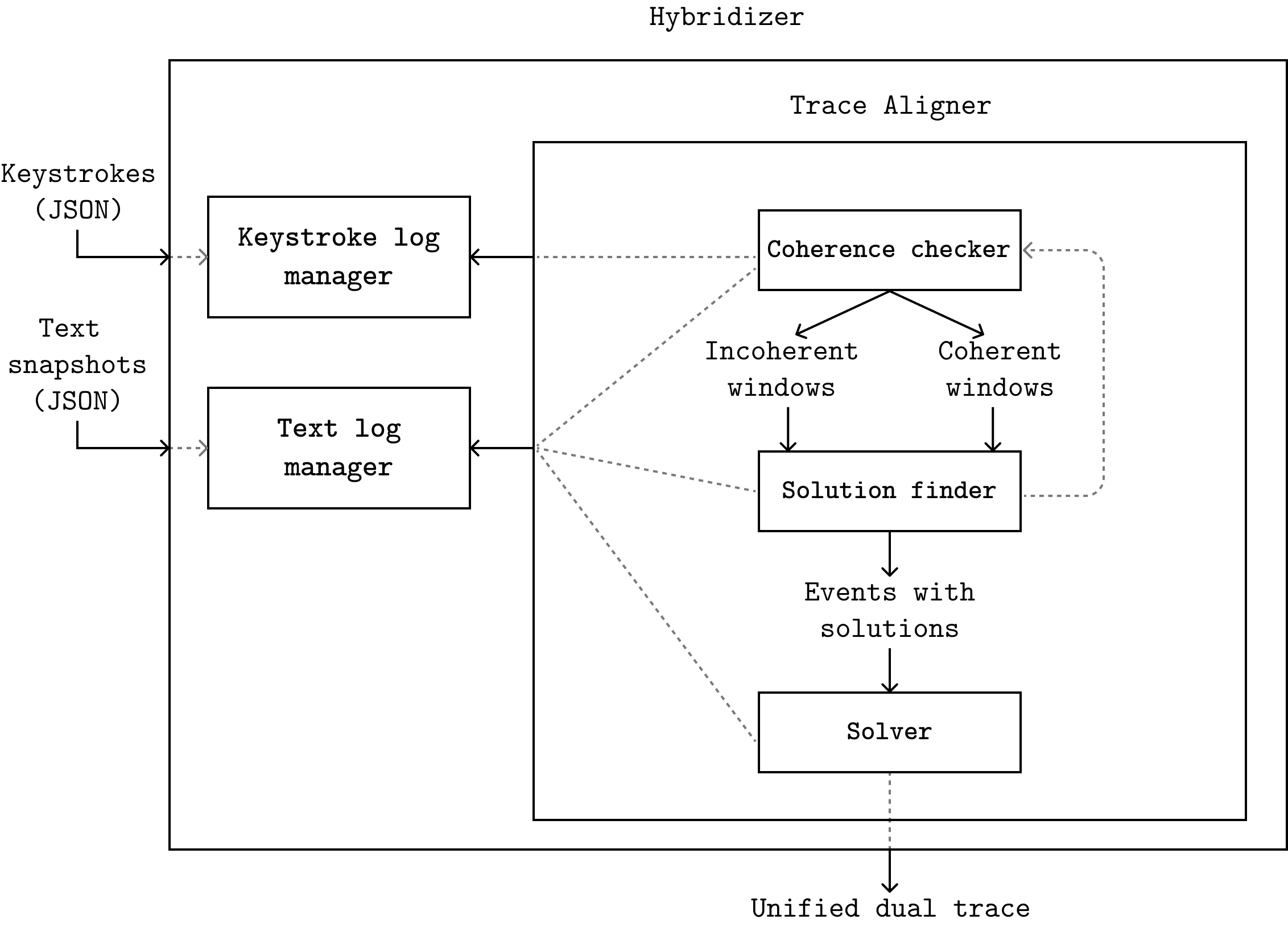}
\caption{The architecture of the hybridizer module \begin{rev} and its core components. 
\end{rev}}
\label{fig:hybridizer-workflow}
\end{figure*}

Hylog's integration is achieved through a \textit{hybridizer} module (Figure \ref{fig:hybridizer-workflow}) with the following architecture: the input logging traces from Inputlog and Hylog, formatted as JSON data, are handled by a \textit{keystroke log manager} and a \textit{text log manager}, which are both queried by a \textit{trace aligner}. The \textit{keystroke log manager} sequentially parses keylogged events, while the \textit{text log manager} reconstructs text states at each logged timestamp as needed. To ensure temporal integrity and mitigate the risk of misordering due to system-level processing delays, all timestamps from both loggers are captured using a high-resolution monotonic clock. The \textit{trace aligner} then aligns keystroke and text-log events by processing the logs through three passes, \textit{coherence checker}, \textit{solution finder}, and \textit{solver}, described below.

\subsubsection{Coherence checker}
In this pass, the \textit{keystroke log manager} incrementally retrieves keystroke events and compares them with the text reconstructed by the \textit{text log manager}, to identify all events where keys pressed at a given time have a direct match with Latin letters that appear on the screen at that same time.
For each keypress, the comparison examines text events up to and including the one whose timestamp immediately 
follows the keypress timestamp. Matched events are \textit{coherent events} and all others are \textit{incoherent events}. A sequence of consecutive coherent (resp. incoherent) events is called a \textit{coherent window} (resp. \textit{incoherent window}).
\begin{rev}
    
\end{rev}

\begin{figure*}[h]
\centering
\includegraphics[width=0.7\textwidth]{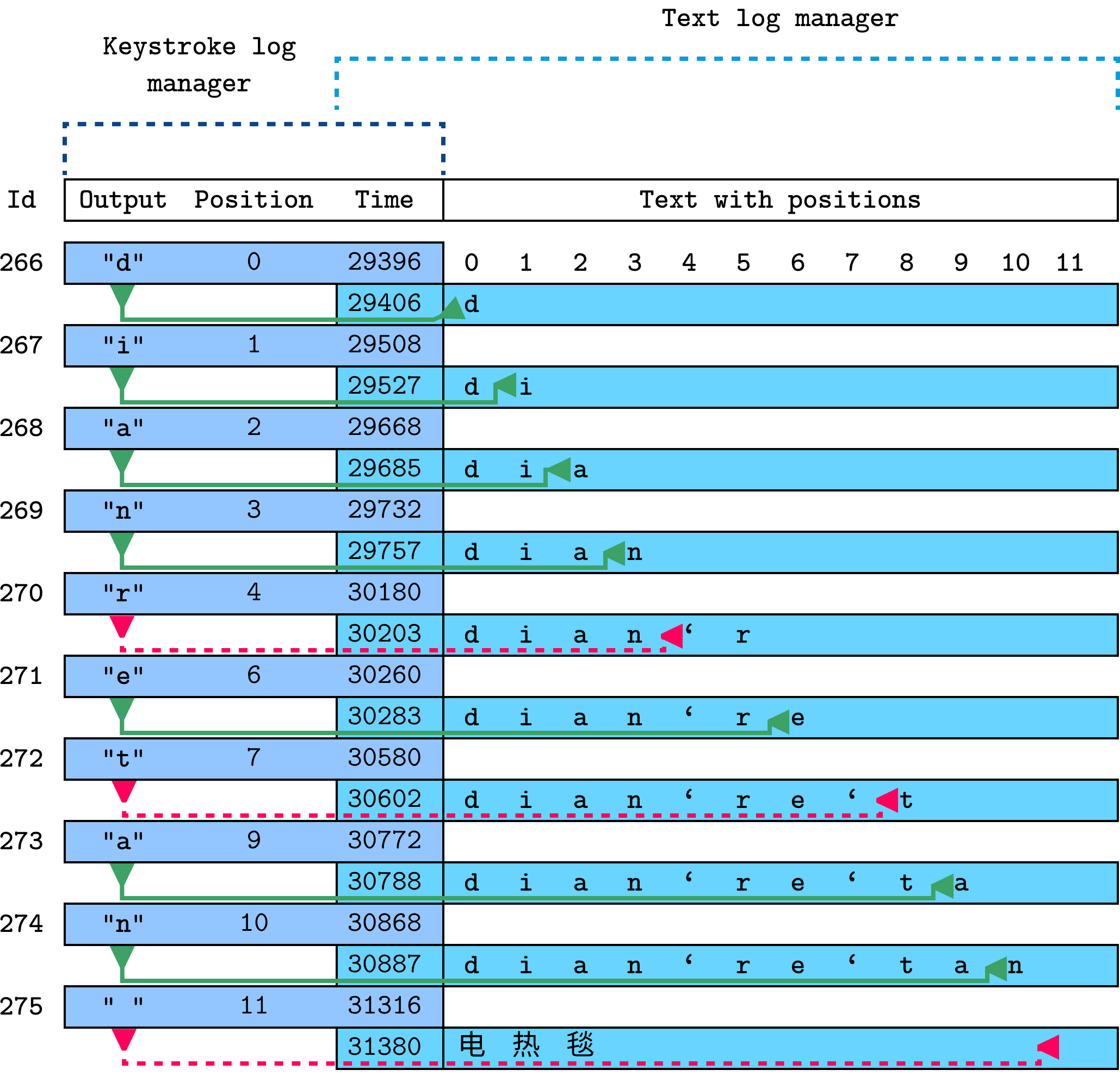}
\caption{Coherence check of an excerpt from a logging session in Chinese. \begin{rev} Coherence-checking stage of the hybridization workflow applied to an excerpt from a Chinese logging session. Green \begin{revv}solid\end{revv} arrows indicate coherent events, red dashed arrows indicate incoherent events requiring further processing with the solution finder module.
\end{rev}}
\label{fig:hybridizer-coherence-check}
\end{figure*}

Figure \ref{fig:hybridizer-coherence-check} illustrates the results of our coherence checking process with an example. In this example, the typist writes the Chinese term for \textit{electric blanket}. She first types the pinyin transcription \textit{dianretan} and then confirms her choice of the non-alphabetic text replacement proposed by the IME system, \begin{CJK}{UTF8}{gbsn}`电热毯'\end{CJK}. Hylog recorded the events 266–275 in the figure, which indicate the typed letters (up to event 274) and the space code (event 275) that corresponds to pressing \texttt{SPACEBAR} to confirm the text replacement. The figure pairs each keystroke event with the text that the \textit{text log manager} incrementally reconstructed from Hylog's logging track. 

Based on the reconstructed texts, the \textit{coherence checker} marks the events 266-269, 271 and 273-274 (highlighted with green solid arrows in the figure) as coherent. That is, for each of these events, a typed letter is in the text at the position indicated by the keystroke log. Hylog's \textit{checker} flags events 270 and 272 as incoherent (red dashed arrows) because the typed letters introduce extra quotes, and event 275 because the space entered with the \texttt{SPACEBAR} key corresponds to Chinese-character input confirmation. Eventually, the results are organized as the following coherence and incoherence windows:

\begin{itemize}
 \item Coherent windows: \texttt{[266, 269]}, \texttt{[271, 271]}, \texttt{[273, 274]}; 
 \item Incoherent windows: \texttt{[270, 270]}, \texttt{[272, 272]}, \texttt{[275, 275]}.
\end{itemize}
 
\subsubsection{Solution finder}
In this pass, Hylog's \textit{solution finder} module resolves the remaining incoherences by cross-checking each incoherent window against pattern-matching rules for known root causes of the mismatches between logged keystrokes and the text reconstructed by Hylog. Each incoherence window is further packaged in a triple set that contains both its preceding and following coherent windows, to foster a broader vision of the context. The first matching pattern (if any) is saved for later use in the \textit{solver} pass (below). 
Table \ref{tab:patterns} describes the rules applied in our Microsoft Word plug-in for Chinese. 

In the above example, windows $[270, 270]$ and $[272, 272]$ contain incoherences that match the syllabic division rule, and the incoherence $[275, 275]$ matches the IME confirmation rule. 

\begin{rev}
The rules in the table aim to demonstrate the feasibility of our approach, but are not intended to be exhaustive. Rather, they serve as a prototype for future improvements. Moreover, they apply specifically for Inputlog 9.5, used with Simplified Chinese and the Microsoft Pinyin IME.\end{rev} Hylog's \textit{solution finder} is designed for extensibility, \begin{rev}as it allows\end{rev} additional rules \begin{rev}to be introduced\end{rev} for other \begin{rev}keyloggers,\end{rev} editors and non-alphabetic scripts.

\begin{table*}[h]
\centering
\caption{Pattern-matching rules of Hylog's \textit{solution finder} in our Microsoft Word plug-in for Chinese.}
\label{tab:patterns}
\begin{tabularx}{\linewidth}{
>{\hsize=0.5\hsize}X
>{\hsize=1\hsize}X
>{\hsize=1.5\hsize}X
}
\toprule
\textbf{Rule name} & \textbf{Incoherence root cause} & \textbf{Rule description} \\
\midrule
Syllabic division &
The Chinese IME automatically inserts a separator \textit{'} in the Pinyin transcription, to visualize syllable separation. &
The Latin letter entered by the user appears immediately after a separator \textit{'} in the text, rather than at the expected position. \\

Syllabic separator deletion &
After \texttt{BACKSPACE} is pressed by the typist, the Chinese IME removes a syllable separator (added previously) from the Pinyin transcription. &
In response to a single \texttt{BACKSPACE} press, the Pinyin transcription recorded by Hylog loses two Latin letters instead of just one. Additionally, the Pinyin transcription recorded before the \texttt{BACKSPACE} contains a syllable separator \textit{'}, immediately preceding the deleted character. \\

Chinese punctuation &
The typist types a Chinese punctuation mark (e.g., \begin{CJK}{UTF8}{gbsn}\textit{。}\end{CJK}) but the keystroke logger records the Latin version (e.g., \textit{.}) of the same mark. &
The keystroke logger records a Roman alphabet punctuation mark entered by the typist in a specific position. At that position, the text logged by Hylog contains the corresponding Chinese version of the mark. \\

IME confirmation &
The typist presses \texttt{SPACEBAR} or a number to confirm a replacement candidate from the IME. &
The keystroke logger records a number key or \texttt{SPACEBAR}. The character corresponding to that key does not appear in the text logged by Hylog at the expected position. \\
\bottomrule
\end{tabularx}
\end{table*}

\begin{rev}
If pattern matching does not succeed,
the solution finder module re-invokes the coherence checker, comparing the incoherent event with the text some instants later than the timestamp of the current event. 
In fact, the text logged by the text logger
may appear with small delay after a keypress logged by the keylogger.

Even if such delays in snapshot acquisition may increase the complexity of the alignment process, such temporal drifts do not affect the timestamps computed by Hylog. Eventually, Hylog will consider the accurate timestamps from the keystroke trace, which precisely correspond to the keyboard events directly originated by the typist.
Although keyloggers do not ensure absolute precision
\begin{revv}
(this may represent a problem for studies that strictly require millisecond-level granularity)\end{revv}, they are the most accurate applications among input loggers.
\end{rev}
 
\subsubsection{Solver}
Finally, Hylog's \textit{solver} module generates the dual trace by rewriting the keystroke log and integrating the corrections identified for each incoherent window. For example, when a syllabic division occurs, the module adjusts character positions in the keystroke log by shifting position values accordingly. For IME confirmations, the log includes both the confirmed Chinese text and its corresponding pinyin transcription, annotated with start and end indices.

\begin{figure*}[h]
\centering
\includegraphics[width=0.7\textwidth]{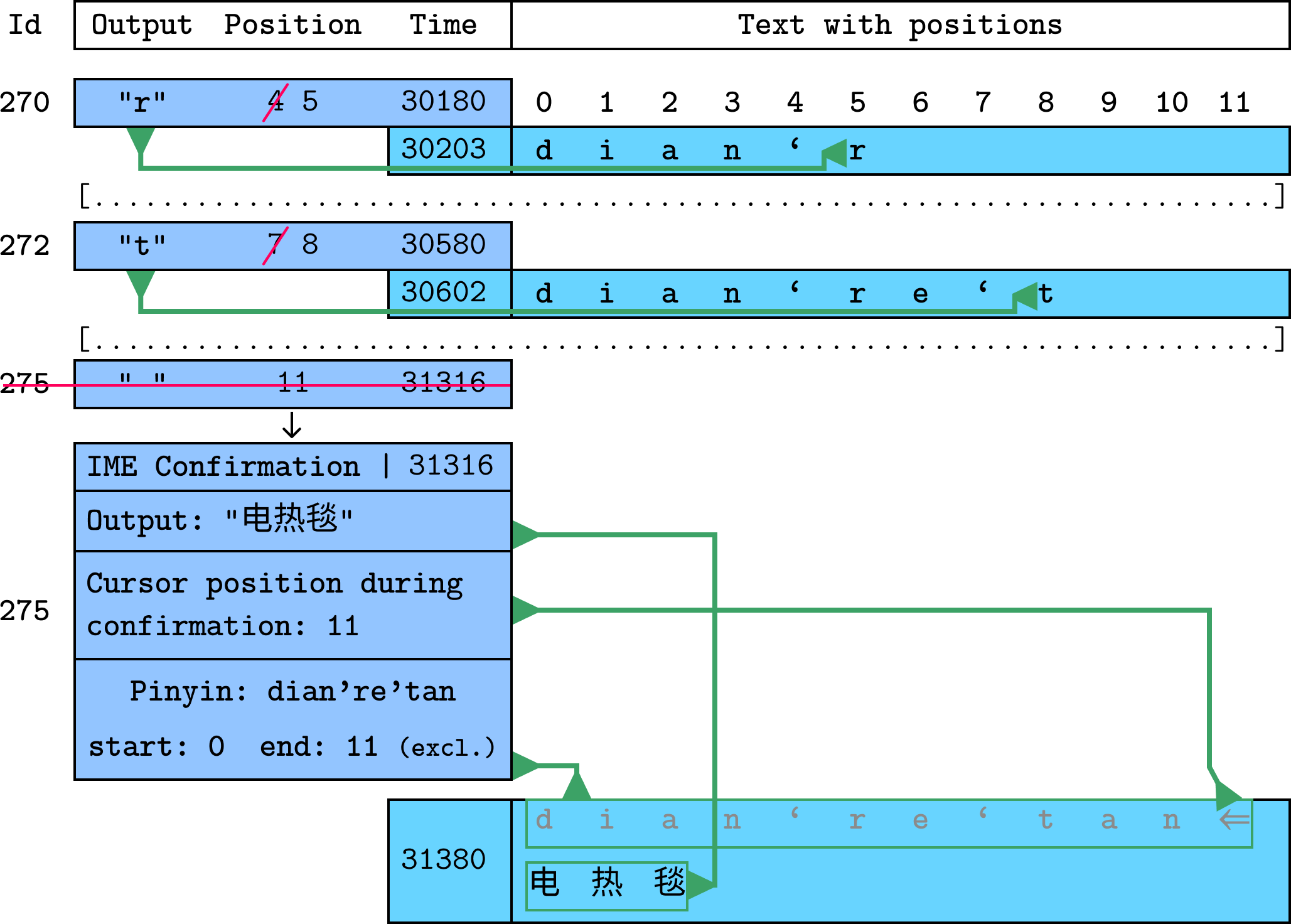}
\caption{Application of the solutions found for three incoherencies \begin{rev} (solver stage). The figure shows how the solutions identified by the solution finder are applied to resolve incoherent events in the keystroke trace.
\end{rev}}
\label{fig:hybridizer-solutions-applier}
\end{figure*}

Figure \ref{fig:hybridizer-solutions-applier} shows the dual trace for the example, highlighting the changes made in the keylog data, after applying the solutions for the incoherent events (270, 272, 275).

This three-pass process is robust enough to handle complex, non-linear editing. For instance, when a typist deletes a confirmed Chinese character to replace it with another, the DSW algorithm captures the change in the text state while the keylogger records the backspace and new pinyin keystrokes. Hylog's hybridizer module then correctly aligns these events, resolving the incoherence to ensure the final dual trace accurately represents the revision before it is passed to the analyzer.

\subsection{Analyzer}\label{subsection:analyzer}

The dual trace produced by Hylog's \textit{hybridizer} serves as input to the \textit{analyzer} module, which computes quantitative metrics describing writing behavior. Hylog's prototype, developed as a proof of concept and reported here, allows measurements about the set of IKIs for Chinese, as discussed in §~\ref{subsection:inputlog-analysis}. Below we outline the \textit{analyzer}'s processing steps for each IME-confirmation event detected in the dual trace. 

\paragraph{Identification of the pinyin syllables} 
For each Chinese string confirmed and inserted, the corresponding pinyin string is segmented based on the syllable separator. Each syllable is treated as an independent processing unit, as illustrated in this example:

\begin{itemize}
 \item \textbf{IME confirmation event}: 289
 \item \textbf{Chinese string}: \begin{CJK}{UTF8}{gbsn}这产品\end{CJK} (`this product')
 \item \textbf{Pinyin string in the log}: `zhe'chan'pin'
 \item \textbf{Pinyin syllables}: [zhe, chan, pin]
\end{itemize}

\begin{figure*}[h]
\centering
\includegraphics[width=0.7\textwidth]{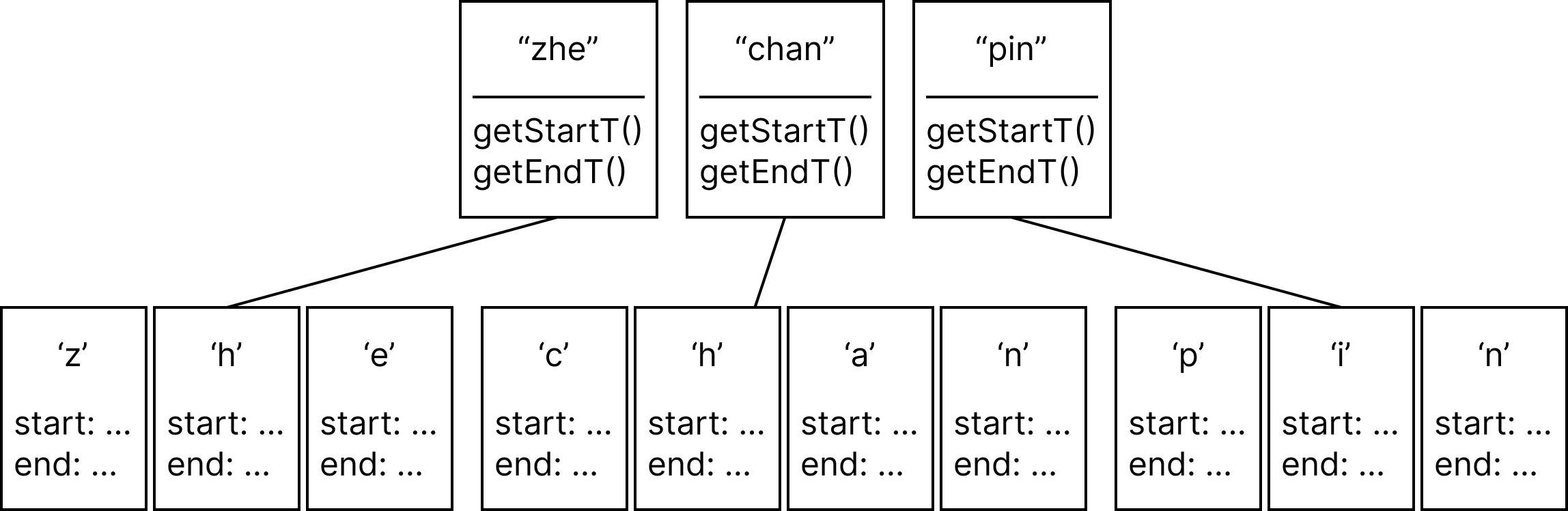}
\caption{Syllables and characters with timestamps  \begin{rev} (analyzer module). The figure shows the propagation of timestamps from Latin characters to the corresponding pinyin syllables.
\end{rev}}
\label{fig:analyzer-characters-segmentation-with-timestamps}
\end{figure*}

\paragraph{Extraction of timestamps for each character}\label{timestamps-extraction-phase} 
Timestamps are collected for every character of the pinyin syllables, and the timestamps of the initial and final Latin letter of each syllable are propagated as start and end timestamps of the corresponding Chinese character (Figure \ref{fig:analyzer-characters-segmentation-with-timestamps}). The dual trace stores monotonic Inputlog timestamps converted to UNIX format with millisecond precision.

\paragraph{Segmentation of Chinese text} As noted in \S~\ref{subsection:inputlog-analysis}, Chinese word boundaries are not marked orthographically, and lexical units vary considerably in length. Automatic segmentation is therefore essential to identify word boundaries and analyze IKIs. The Chinese text entered by the confirmation event in the log is segmented into words with the \textit{Stanford CoreNLP Chinese Segmenter} \cite{tseng2005} (Stanford NLP Group, version 4.5.9), together with the corresponding Chinese models package (version 4.5.9). The segmenter is configured with the \textit{ctb.gz} model (Chinese Treebank), the serialized dictionary \textit{dict-chris6.ser.gz}, and SIGHAN-based post-processing. The CoreNLP pipeline includes the tokenize and ssplit annotators. For instance:

\begin{itemize}
\item \textbf{Chinese string}: \begin{CJK}{UTF8}{gbsn}这产品\end{CJK};
\item \textbf{Chinese words from segmenter}: [\begin{CJK}{UTF8}{gbsn}这, 产品\end{CJK}].
\end{itemize}

\begin{figure*}[h]
\centering
\includegraphics[width=0.7\textwidth]{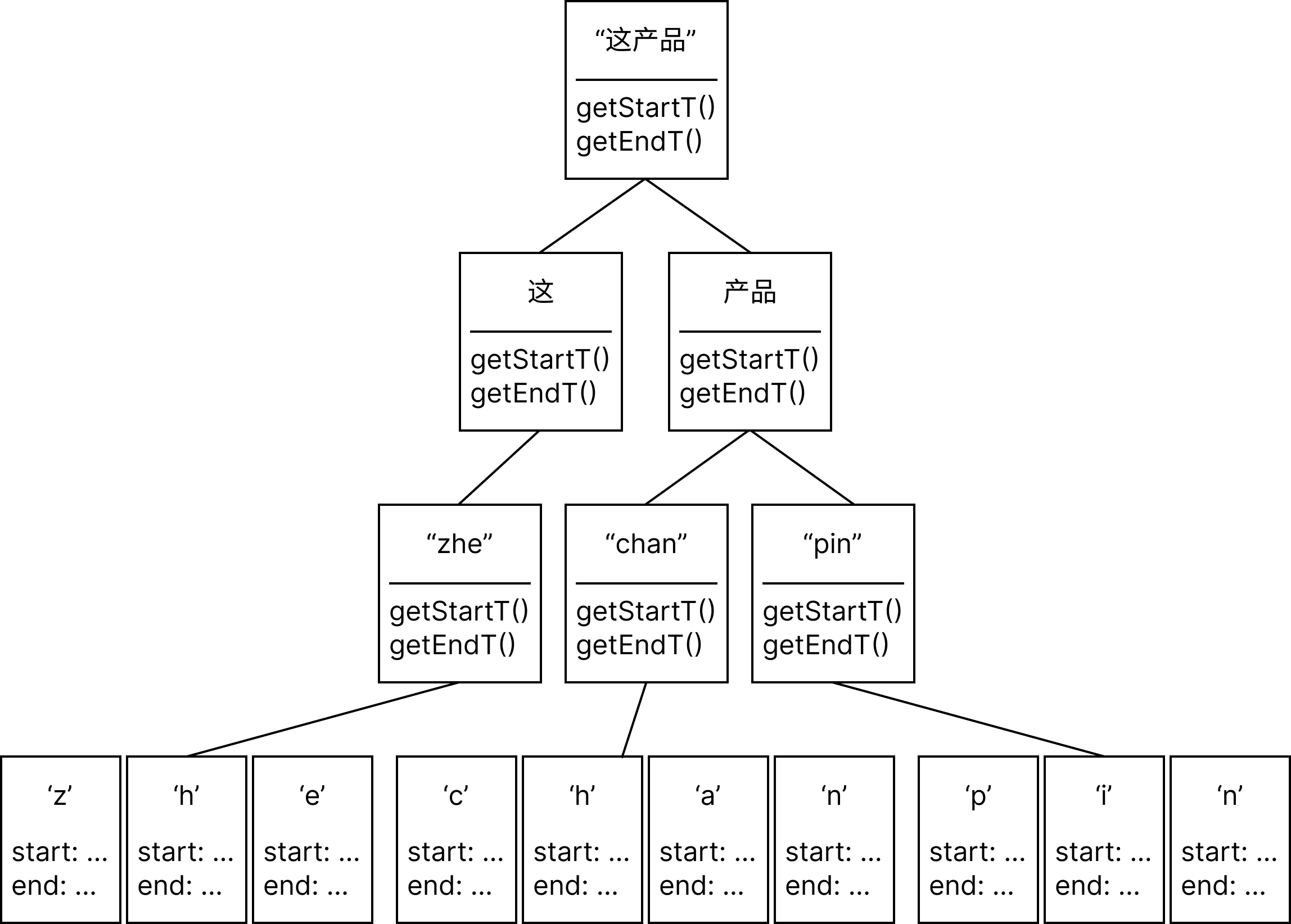}
\caption{Full segmentation tree for the Chinese string \begin{CJK}{UTF8}{gbsn}`这产品'\end{CJK} \begin{rev}(analyzer module). The figure shows the hierarchical segmentation tree for the Chinese string \begin{CJK}{UTF8}{gbsn}`这产品'\end{CJK} that illustrates the relationships between the text with logograms, Chinese words, pinyin syllables, and Latin characters.
\end{rev}}
\label{fig:analyzer-segmentation-tree}
\end{figure*}

\paragraph{Computation of the metrics} 
All collected text strings are organized in an inverted tree, propagating timings from the leaves (characters) up to the root (Chinese text). Figure \ref{fig:analyzer-segmentation-tree} depicts the hierarchical tree structure generated for this example. 

The tree representation is used to compute the following metrics:

\begin{itemize}
 \item[$\wwlatinchar$] \textbf{IKIs between Latin letters}:
 \begin{enumerate*}[label=(\roman*)]
 \item the list of Chinese words is retrieved (level 1 in the tree);
 \item for each word, the list of pinyin syllables is retrieved (level 2);
 \item for each syllable, the list of characters is retrieved (level 3) and set in ascending order, following start timestamps;
 \item an IKI between two characters in the list is \\ $currentChar.start -previousChar.start$.
 \end{enumerate*} 
 
 \item[$\wwpinyinsyll$] \textbf{IKIs between pinyin syllables or characters}:
 \begin{enumerate*}[label=(\roman*)]
 \item the list of Chinese words is retrieved (level 1);
 \item for each word, the list of pinyin syllables is retrieved (level 2) and set in ascending order, following start timestamps;
 \item an IKI between two syllables in the list is \\ $current.getStartT() -previous.getEndT()$.
 \end{enumerate*}
 
 \item[$\bwpinyin$] \textbf{IKIs between words}:
 \begin{enumerate*}[label=(\roman*)]
 \item the list of Chinese words is retrieved (level 1) and set in ascending order;
 \item an IKI between two words in the list is $current.getStartT() -previous.getEndT()$.
 \end{enumerate*}
 
 \item[$\imeconf$] \textbf{IKIs before IME confirmation}:
 \begin{enumerate*}[label=(\roman*)]
 \item the list of Chinese words is retrieved (level 1) and set in ascending order, following start timestamps;
 \item the first Chinese word is selected from the list;
 \item the IKI between the word at Chinese level is the IKI between the word extracted from the list and the previous IME confirmation event \\ $word.getStartT()- \\previous\_ime\_confirmation.getStartT()$.
 \end{enumerate*}
 
 \item[$\bwchinese$] \textbf{IKIs after IME confirmation}: considering the Chinese string introduced, the IKI is \\$ime\_confirmation.getStartT() - text.getStartT()$.
\end{itemize}

Having described Hylog's complete architecture---from ecological text logging through hybridization to metric computation---we now evaluate whether this system can reliably capture these measurements under realistic text-production conditions.

%% file: 4-empirical-testing.tex
\section{System validation}\label{sec:system-validation}
We evaluated Hylog in a case study primarily designed as a proof-of-concept to validate Hylog's technical reliability, synchronization accuracy, and internal consistency under realistic text-production conditions.
A secondary aim was to generate illustrative examples of the fine-grained analytical insights described in \S~\ref{subsection:analyzer}: IKIs between Latin letters within pinyin syllables, characters, and words; and IKIs surrounding IME confirmation events.

We compared these results with those computed using only Inputlog 9.5 for alphabetical languages, which is still unaware of Chinese-level structure or IME events. Larger discrepancies indicate larger potential errors when relying on an alphabet-only keylogger and, correspondingly, greater added value of our hybrid approach. Given the small sample (two  participants), results are only illustrative. 

The participants were two experienced translators, here pseudonymized as \textit{John} and \textit{Jane}\footnote{Gender was not a variable in the study, so names are random and do not necessarily match the participants' genders.}. Both provided verbal informed consent prior to participation.
\begin{revv}
Approval from an ethics committee was not deemed necessary, as the study involved a non-invasive task, collected no sensitive personal data, and posed no foreseeable risk to the participants.
\end{revv}
John, a postdoctoral researcher and interpreting instructor, was a Chinese L1 (``native'') speaker. Jane, a PhD student, was a Chinese L2 speaker (L1 Italian). We deliberately selected two markedly different profiles to introduce variation and test the robustness of our approach across differing degrees of task experience and input fluency. They were tasked with translating a one-page heating-pad instruction leaflet from English into Chinese and we logged their activity during the entire session with both Inputlog and Hylog.

They used Microsoft Word to produce the Chinese translation and Google Chrome to consult online sources and dictionaries—this reflects typical translator workflows. We installed our Hylog plug-ins for Word and Chrome and also recorded screens with Open Broadcaster Software to enable later visual cross-checking. 
Both sessions were conducted on Windows 10 Enterprise LTSC (x64) systems. John's computer ran version 22H2 (build 19045.5608) on an Intel Core i7-8700 @ 3.20 GHz with 16 GB RAM, and Jane's ran version 1809 (build 17763.7136) on an AMD Ryzen 3 PRO 3200GE @ 3.30 GHz with 8 GB RAM. For logistical reasons, we used Microsoft Pinyin IME even though Sogou Pinyin accounts for most Chinese input ($\approx$80 per cent). Microsoft's IME remains common in institutional setups because it comes bundled with Windows, as on all department computers. John's system used version 10.0.19041 branch, and Jane's the legacy 10.0.17763 branch. 
We validated the coherence checker, solution finder, and text reconstruction accuracy during the hybridization process (operated by the text log manager) by visually comparing their outputs with a small portion of video-verified data.

The two logged sessions differed in length, event density, and output size. For John (L1 Chinese), we logged \begin{rev}3,155 events related to activity on the Word document and within Chrome,\end{rev} in a session lasting 37 minutes and 57 seconds. His final Chinese translation contained 430 logograms, forming 280 words (as per the Stanford segmenter), reflecting the typical one-to-two-character composition of Chinese words. His mean lag between a keypress event and the logging of the corresponding rendered text was 39,76 ms for Word plug-in and 172,77 ms for Chrome plug-in. 
For Jane (L2 Chinese) we logged \begin{rev}2,566 events (Word and Chrome activity)\end{rev}
in a session lasting 33 minutes and 17 seconds. Her final translation contained 186 logograms forming 117 words. Her mean lag between key presses and the corresponding text being logged by the text logger is 138,39 ms for Word plug-in and 11,12 ms for Chrome plug-in.

\begin{rev}
\subsection{Statistics on hybridization performance}
Table \ref{tab:hyb:stats:john:and:jane} reports the impact of the hybridization algorithm for the log data in our case study.
In the case of John, the hybridization algorithm reduced the number of incoherent events from 758 (24\% of the total) to 93 (2.9\% of the total), meaning that 88\% of the inconsistencies have been resolved. For Jane, the reduction was from 907  incoherent events (35\% or the total) to 243 (9.5\% of the total), that is, the hybridization algorithm resolved 73\% of the inconsistencies. 

\begin{table}
\centering
\begin{rev}
\centering
\caption{Traces alignment statistics for John's and Jane's session.}
\begin{tabular}{lcc} 
\toprule
 & \begin{tabular}[c]{@{}c@{}}\textbf{John }\end{tabular} & \begin{tabular}[c]{@{}c@{}}\textbf{Jane}\end{tabular} \\ 
\midrule
\begin{tabular}[c]{@{}c@{}}\textbf{\textbf{No. events}}\end{tabular} & 3,155 & 2,566 \\ 
\textbf{No. incoherent events} & 758 (24\%) & 907 (35\%) \\
\begin{tabular}[c]{@{}l@{}}\textbf{No. incoherent events}\\\textbf{after hybridization}\end{tabular} & 93 (2,9\%) & 243 (9,5\%) \\ 
\bottomrule
\end{tabular}\label{tab:hyb:stats:john:and:jane}
\end{rev}
\end{table}

Table \ref{tab:hybridization-results} indicates how the remaining inconsistencies impact on the results of our approach, with respect to the ability of precisely identifying IME confirmations.\end{rev} For John's session, Hylog identified 305 IME confirmations, 300 of these being actual confirmations, while 5 of them being false positives---i.e., Inputlog's key events not corresponding to genuine IME confirmations---over a total of 301 pinyin strings actually confirmed (as counted from the screen recording) by the informant. Hylog incurred only one false negative, \begin{rev}
due to yet-unresolved incoherent events.
Thus, it achieved 98,4\% precision and 99,6\% recall. \end{rev}For Jane's session, the Hylog identified 82 IME confirmations over a total of 86 pinyin strings actually confirmed (as counted from the screen recording), with no false positives and a total of 4 false negatives due to yet-unresolved incoherent events, \begin{rev}with precision and recall of 100\% and 95,6\%, respectively.\end{rev}

\begin{rev}
\begin{table}
\centering
\begin{rev}
\caption{Analytical data about accuracy and completeness of the hybridization process for John's and Jane's sessions.}
\arrayrulecolor{black}
\begin{tabular}{lll} 
\toprule
 & \textbf{John} & \textbf{Jane} \\ 
\arrayrulecolor{black}\midrule
\textbf{No. IME confirmations submitted} & 301 & 86 \\
\textbf{No. IME confirmations detected with Hylog} & 305 & 82 \\
\textbf{No. spurious detections (false positives)~} & 5 & 0 \\
\textbf{No. missed detections (false negatives)} & 1 & 4 \\
\midrule
\textbf{Precision} & 98,4\% & 100\% \\
\textbf{Recall} & 99,6\% & 95,6\% \\
\bottomrule
\end{tabular}\label{tab:hybridization-results}
\end{rev}
\end{table}
\end{rev}

\subsection{Language-unit-based analysis}\label{sec:language-unit-based-analysis}

Language-based analyses focus on linguistic units (letters, words, syntactic boundaries) and examine interkeystroke intervals (IKIs).

Tables \ref{tab:john-keydown} and \ref{tab:jane-keydown} report the keydown--keydown interval \textit{(press latency)} statistics computed by default by the in-built standard pause analysis of Inputlog 9.5 (I) and we compare them to those by Hylog (H). The tables list averages, medians, and standard deviations (SD) of the latencies both before and after filtering, i.e., before and after discarding the outliers. The criterion to discard them was just single pass ±2 SD, to eliminate hardware or logging artifacts but preserve the natural variation in typing behavior.

The five column groups display distinct linguistic or operational levels: (1) total (all events recorded in the session), (2) Latin letters (keystrokes used for pinyin input), (3) characters (Chinese logograms generated from pinyin syllables), (4) words (lexical units segmented from the final Chinese text), and (5) IME (intervals linked to confirmation or conversion events in the IME). 

\begin{table*}
\centering
\caption{John's keydown--keydown intervals as reported by Inputlog (I) and the hybrid keylogger (H).}
\label{tab:john-keydown}
\begin{tabular}{lcccccccccc}
\toprule
 & \multicolumn{2}{c}{\textbf{Total}} & \multicolumn{2}{c}{\textbf{Letters}} & \multicolumn{2}{c}{\textbf{Characters}} & \multicolumn{2}{c}{\textbf{Words}} & \multicolumn{2}{c}{\textbf{IME}} \\
\cline{2-11}
 & \textbf{I} & \textbf{H} & \textbf{I} & \textbf{H} & \textbf{I} & \textbf{H} & \textbf{I} & \textbf{H} & \textbf{I} & \textbf{H} \\
\midrule
\textbf{Average} & -- & 1496.06 & 515 & 207.54 & -- & 219.84 & 885 & 234.88 & -- & 934.48 \\
\textbf{Median} & -- & 144 & -- & 144 & -- & 88 & -- & 96 & -- & 556.5 \\
\textbf{SD} & -- & 14356.97 & 878 & 685.41 & -- & 1574.99 & 0 & 1925.57 & -- & 2822.05 \\
\textbf{Outlier count} & -- & 11 & -- & 7 & -- & 3 & -- & 1 & -- & 4 \\
\textbf{Outlier \%} & -- & 0.72 & -- & 0.96 & -- & 0.59 & -- & 0.33 & -- & 1.41 \\
\midrule
\textbf{Average (filtered)} & -- & 854.05 & -- & 164.16 & -- & 125.60 & -- & 125.28 & -- & 635.05 \\
\textbf{Median (filtered)} & -- & 143 & -- & 144 & -- & 88 & -- & 96 & -- & 552.5 \\
\textbf{SD (filtered)} & -- & 2753.94 & -- & 118.73 & -- & 179.50 & -- & 179.37 & -- & 509.60 \\
\bottomrule
\end{tabular}
\end{table*}

\begin{table*}
\centering
\caption{Jane's keydown--keydown intervals as reported by Inputlog (I) and the hybrid keylogger (H).}
\label{tab:jane-keydown}
\begin{tabular}{lcccccccccc}
\toprule
 & \multicolumn{2}{c}{\textbf{Total}} & \multicolumn{2}{c}{\textbf{Letters}} & \multicolumn{2}{c}{\textbf{Characters}} & \multicolumn{2}{c}{\textbf{Words}} & \multicolumn{2}{c}{\textbf{IME}} \\
\cline{2-11}
 & \textbf{I} & \textbf{H} & \textbf{I} & \textbf{H} & \textbf{I} & \textbf{H} & \textbf{I} & \textbf{H} & \textbf{I} & \textbf{H} \\
\midrule
\textbf{Average} & -- & 4340.97 & 531 & 391.05 & -- & 1006.69 & 976 & 154.94 & -- & 1433.46 \\
\textbf{Median} & -- & 145 & -- & 143 & -- & 113 & -- & 112 & -- & 633 \\
\textbf{SD} & -- & 19613.98 & 491 & 866.46 & -- & 6834.24 & 825 & 224.44 & -- & 1833.32 \\
\textbf{Outlier count} & -- & 12 & -- & 10 & -- & 3 & -- & 2 & -- & 4 \\
\textbf{Outlier \%} & -- & 2.61 & -- & 4.41 & -- & 1.91 & -- & 2.38 & -- & 5.00 \\
\midrule
\textbf{Average (filtered)} & -- & 1619.83 & -- & 227.02 & -- & 179.16 & -- & 123.01 & -- & 1136.47 \\
\textbf{Median (filtered)} & -- & 144 & -- & 128 & -- & 113 & -- & 111 & -- & 624.5 \\
\textbf{SD (filtered)} & -- & 5422.88 & -- & 316.64 & -- & 320.23 & -- & 48.90 & -- & 1221.75 \\
\bottomrule
\end{tabular}
\end{table*}

Inputlog 9.5 accurately logs keystrokes for alphabetic languages but cannot access IME state information. That is, it correctly records key events for Latin letters but cannot distinguish the subsequent Chinese character-generation or confirmation steps. These unrecorded processes appear as missing values (–) in columns I of Tables \ref{tab:john-keydown} and \ref{tab:jane-keydown}. By contrast, Hylog allows IKIs to be separated from the keypress latency and computed continuously from letter entry through pinyin composition, character confirmation, and word insertion, and it further integrates IME output. The resulting disparity is instructive. At the smallest sublexical level (letters), Inputlog and Hylog yield broadly comparable values—around 500 ms versus 200–400 ms, depending on participant and filtering---thereby confirming that both systems capture motor-level timing reliably.

The divergence comes in the non-alphabetic columns: Hylog identifies additional peaks in the characters and IME events, with mean latencies near 0.6–1.1 s.
Because Inputlog cannot register those events, its \textit{total} column merges their key presses with their flanking intervals, thereby smoothing away the very temporal patterns that define non-alphabetic input.

Finally, these observations clarify a broader methodological point. As mentioned, Inputlog's ready-made pause analysis module---like most tools of its generation---computes events as keydown--keydown intervals, effectively combining keyhold duration and inter key latency into a single measure. Hylog, by contrast, distinguishes between \textit{dwell time} (keydown--keyup) and genuine \textit{interkeystroke intervals} (keyup--keydown). This distinction enables a finer temporal decomposition of mixed-script activity that becomes crucial when input involves multi-stage events such as character conversion or IME confirmation, which have no mechanical equivalent in alphabetic typing.

The difference also matters empirically. Without IME-state data, Inputlog classifies both translators' sessions as broadly similar, masking the behavioral asymmetry that becomes visible once the IME and character stages are separately logged.

\section{Discussion on Using Hylog for Behavioral analysis}\label{sec:hylog:for:behavioral:analysis}

\begin{rev}

Behavioral approaches analyze IKIs surrounding specific translation or revision events, like typos and searches~\cite{ImmonenMakisalo2010,Immonen2011}.
In what follows, we apply this approach to the non-alphabetic input data of our case study.
We are aware that our dataset is not sufficiently large to draw generalizable conclusions about translators' cognitive processes. \begin{revv} We also point out that our tool recorded text production for only a single non-alphabetic language (Simplified Chinese), involving a specific IME (Microsoft IME). For these reasons,\end{revv} the aim of this section is to showcase the kinds of insights researchers can derive by using Hylog, rather than 
making claims on the cognitive interpretations thereby. \begin{revv}In other terms, the discussion that follows is intended to guide the reader through an illustrative process of hypothesis formulation and not to formally test those hypotheses.\end{revv}

\end{rev}

Hylog's core methodological contribution is decomposing what many prior loggers treat as a single interval measure into three complementary metrics: \begin{enumerate*}[label=(\arabic*)]
\item \textit{dwell times} (keydown--keyup), capturing motor execution; 
\item \textit{positive interkeystroke intervals} (keyup--keydown gaps), capturing situational and attentional variation;
\item \textit{rollovers} or negative interkeystroke intervals (keyup--keydown overlaps), capturing routine entrenchment and automation.
\end{enumerate*}

This tripartite decomposition \begin{rev}
    may allow for
disentangling 
\end{rev}motor and cognitive timing with a precision unusual in customary keylogging for cognitive research, especially in mixed-script input, where alphabetic keystrokes, logogram entries, and IME-driven confirmation events coexist within a single compositional stream. \begin{rev}As a possible example, we now discuss how one could\end{rev} integrate the temporal data from both John and Jane, after outlier removal, \begin{rev}aiming\end{rev} to illustrate the \textit{types} of fine-grained behavioral insights Hylog \begin{rev}might provide\end{rev} regarding execution regularity, attentional variations, and cognitive effort across linguistic levels and user profiles.

\begin{table*}[]
\centering
\caption{Dwell times between processing levels, filtered data (milliseconds).}
\label{tab:dwell}
\begin{tabular}{lrrrrrr}
\toprule
& \multicolumn{3}{c}{\textbf{John (L1 Chinese)}} & \multicolumn{3}{c}{\textbf{Jane (L2 Chinese)}} \\
\cline{2-7}
\textbf{Between units} & \textbf{Mean} & \textbf{Median} & \textbf{SD} & \textbf{Mean} & \textbf{Median} & \textbf{SD} \\
\midrule
\textbf{Latin letters} & 110.95 & 112 & 34.23 & 87.74 & 80 & 21.21 \\
\textbf{Chinese characters} & 98.75 & 96 & 23.79 & 78.49 & 80 & 15.90 \\
\textbf{Chinese words} & 105.29 & 104 & 23.06 & 77.62 & 79 & 15.02 \\
\bottomrule
\end{tabular}
\end{table*}

Dwell times reported in Table \ref{tab:dwell} appear to be a stable motor constant, rather than a site of cognitively driven variation. Both participants maintain remarkably consistent keyhold durations across processing levels (and throughout the session). This \begin{rev} may\end{rev} suggest that pressing and releasing a key is a rehearsed and entrenched motor skill that is not seriously impacted by linguistic complexity or situational circumstances.  Low standard deviations (15--34 ms) show minimal within-subject dispersion. 

Jane's shorter keyholds (median \textasciitilde80 ms) point to a fast typist while John's slightly longer keyholds (means $\approx$ 99--111 ms) reflect a more segmented and typical rhythm \cite{Dhakal2018}. Jane's low standard deviation (15 ms) \begin{rev} may\end{rev} indicate highly consistent motor execution. Mechanical constancy prevails regardless of script or unit. In fact, when both translators switch to logographic mode, their dwell profiles converge. This \begin{rev} may \end{rev} suggest that within-word motor coordination stabilizes at the lexical level, largely independent of proficiency. The methodological implication \begin{rev}could be\end{rev}: motor steadiness can  factored out, so that cognitive aspects can be studied directly in the interkeystroke intervals without distortions. 

\begin{table*}[]
\centering
\caption{Positive interkeystroke intervals between language units, filtered data (milliseconds).}
\label{tab:positive_iki}
\begin{tabular}{lrrrrrr}
\toprule
& \multicolumn{3}{c}{\textbf{John}} & \multicolumn{3}{c}{\textbf{Jane}} \\
\cline{2-7}
\textbf{Between units} & \textbf{Mean} & \textbf{Median} & \textbf{SD} & \textbf{Mean} & \textbf{Median} & \textbf{SD} \\
\midrule
\textbf{Latin letters} & 104.66 & 80 & 134.91 & 169.92 & 64 & 333.84 \\
\textbf{Chinese characters} & 169.57 & 80 & 462.95 & 268.88 & 64 & 1493.05 \\
\textbf{Chinese words} & 129.47 & 88 & 292.37 & 59.75 & 63 & 46.94 \\
\textbf{IME} & 513.22 & 424 & 503.94 & 1051.55 & 512.50 & 1218.84 \\
\bottomrule
\end{tabular}
\end{table*}

Hylog's data support nuanced analyses of positive IKIs to better understand the cognitive aspects of typing behavior. Some statistics are presented in Table \ref{tab:positive_iki}. Our text logger successfully captures their right-skewed distributions, which cluster around short transitions but include occasional long gaps. This \begin{rev}may\end{rev} provide researchers with  data to investigate whether these gaps correspond to activities such as reading, planning, or decision-making. For example, Hylog's output shows that intervals lengthen from letters to characters, then shorten again at the word level. This pattern enables researchers to formulate and test hypotheses about how cognitive effort might fluctuate within the linguistic hierarchy. Longer inter-character IKIs might respond to the momentary attentional demand of deciding whether adjacent characters form one word or belong to separate ones (see footnote in \S~\ref{introduction} and below). Longer inter-word IKIs may, in turn, result from chunking and smoother coordination within established language units that can also occur in isolation.

The only sharp increase occurs during IME confirmations, when visual scanning, selection, and outcome checking intervene. This surge may reflect a distinct cognitive bottleneck unrelated to motor control and \begin{rev}suggests\end{rev} that Hylog \begin{rev}can distinguish\end{rev} interface-driven processes within the cognitive flow of writing from typing mechanics. Conventional keydown-based measures would collapse all this into a single latency value and miss or mask the underlying variations in cognitive effort.

\begin{table*}[]
\centering
\caption{Rollover (negative IKIs) between language units, filtered data.}
\label{tab:rollover}
\begin{tabular}{llrrrrr}
\toprule
\textbf{Participant} & \textbf{Between units} & \textbf{Mean} & \textbf{Median} & \textbf{SD} & \textbf{Count} & \textbf{\%} \\
\midrule
\textbf{John} & \textbf{Latin letters} & -43.24 & -32.50 & 24.38 & 193 & 26.44 \\
 & \textbf{Chinese characters} & -50.53 & -48 & 24.09 & 348 & 68.91 \\
 & \textbf{Chinese words} & -55.43 & -55 & 27.21 & 202 & 66.01 \\
\midrule
\textbf{Jane} & \textbf{Latin letters} & -28.30 & -17 & 16.40 & 21 & 9.25 \\
 & \textbf{Chinese characters} & -23.83 & -16 & 11.15 & 38 & 24.36 \\
 & \textbf{Chinese words} & -21.39 & -16 & 9.38 & 20 & 23.81 \\
\bottomrule
\end{tabular}
\end{table*}

Rollover IKIs in Table \ref{tab:rollover} provide a complementary index of motor fluency. John's keystrokes show extensive overlaps, with means ranging from --43 to --55 ms. Jane's corresponding overlaps are far lower and point to a more serial execution style. The contrast outlines two control architectures: John's fluent, overlapping rhythm, characteristic of entrenched motor chaining, and Jane's deliberate, segmented coordination. Such negative intervals, often invisible in keydown--keydown measures, quantify the degree of parallelization and directly measure proceduralized skill.

This motor-fluency contrast also interacts with linguistic factors in Chinese typing, as pinyin does not always reflect how Mandarin is actually spoken. For instance, the standard pinyin transcription for \begin{CJK}{UTF8}{gbsn}`只'\end{CJK} (`only') is \textit{zhi}, but speakers often pronounce it closer to [jer], which might be miswritten or misinterpreted as \textit{zhe} (\begin{CJK}{UTF8}{gbsn}`这'\end{CJK}, `this')---just as English speakers hear [shee] (\begin{CJK}{UTF8}{gbsn}`西'\end{CJK}, `west') and would think of typing \textit{shi} rather than \textit{xi}. Besides, regional differences are significant: speakers who merge [n] and [l]---as they do in southwestern regions like Sichuan and Yunnan---might type \textit{lan} for both \begin{CJK}{UTF8}{gbsn}`南'\end{CJK} (\textit{nán}, `south') and \begin{CJK}{UTF8}{gbsn}`蓝'\end{CJK} (\textit{lán}, `blue'), mixing two different words.

Furthermore, about 30 per cent of China's population speaks Cantonese, Wu, Min, or other Chinese languages as L1, yet shares the same written characters: a Cantonese speaker sees \begin{CJK}{UTF8}{gbsn}`人'\end{CJK} (\textit{rén}, `person') and thinks [jan], while a Hokkien speaker thinks [lang] for the same character. Chinese typists thus often need to reconcile a double graphic representation (the target character vs. the pinyin string) and a double phonetic representation (the user's pronunciation vs. standardized pinyin). This conflict is significant not only for L2 learners but also for Mandarin L1 speakers. IMEs usually correct such variations to standard pinyin, but that very correction makes the problem endemic.

Our data suggest that typing skill is not a straightforward matter that can be accounted for with an overall indicator for typing speed. For instance, while John might be assumed to be the faster typist overall, it is Jane who exhibits superior motor speed. Jane's dwell times and sub-lexical IKIs are consistently faster than John's, indicating a higher mechanical fluency when executing pinyin keystroke sequences. Yet, this narrative is inverted when we isolate the cognitive task of IME confirmation. Here, John shows a clear advantage. His median IKI before an IME event is 424 ms, while Jane's is significantly longer at 512.50 ms. The difference is even more pronounced in the mean values (513.22 ms for John vs. 1051.55 ms for Jane). 

In brief, the data for Jane show frequent, much longer IKIs during character selection. This highlights Hylog's power to generate data that can help distinguish between motor-level fluency and language-specific cognitive bottlenecks, such as those seemingly related to recognizing and selecting the correct character. The rollover data (Table \ref{tab:rollover}) further supports this distinction between motor speed and typing flow. This nuanced distinction---between raw motor speed and cognitive processing efficiency---is precisely the kind of hypothesis that traditional keyloggers obscure and that Hylog is designed to enable researchers to investigate.

Together, the three temporal layers sketch a coherent control architecture. Dwell time indexes mechanical stability; positive IKIs capture cognitive variations; rollover traces motor synchronization. John's profile---steady keyholds, short flight times, deep overlap---reflects entrenched fluency. Jane's---equally stable keyholds but longer, more variable IKIs and minimal overlap---exemplifies intentional segmentation and lower automatization. At the IME level, both profiles converge in exhibiting large, irregular IKIs. This \begin{rev}suggests\end{rev} that candidate confirmation, rather than execution, accounts for the decisive temporal cost of Chinese input.

By decomposing timing into three components, Hylog links temporal variance to its functional source and shows how experience, script type, and input method jointly shape the temporal dynamics of text production. The approach achieves fuller linguistic segmentation in non-alphabetic input while producing empirically interpretable, behaviorally grounded \begin{rev}insights\end{rev} on typing automation and cognitive dynamics.

%% file: 5-discussion-related-work-conclusions.tex
\section{Conclusions}
\label{sec:conclusions}

Our analysis of Inputlog 9.5 highlights the inherent limitations of keyloggers designed for alphabetic scripts when confronted with IME-mediated input—a challenge that has motivated both Hylog and the newer, language-specific Inputlog releases. According to the Inputlog documentation, detailed analyses in these new versions require post-processing the general log, whereas Hylog \begin{revv}embodies an\end{revv}
IME-agnostic architectural solution by synchronizing raw keystrokes with the rendered text from the outset. This design yields three main advantages:

\begin{enumerate}
    \item \textbf{Generality and flexibility}. By aligning keyboard output with the final text, Hylog remains robust across IMEs and allows participants to use their preferred input tools, thereby enhancing ecological validity.
    \item \textbf{Extensibility}. Its dual-trace architecture provides a model for synchronizing multiple data streams—such as keystrokes, text, screen recording, or EEG—facilitating multimodal analyses of complex activities like revision or post-editing.
    \item \textbf{Transparency and openness}. Both the DSW and hybridizer algorithms are fully specified, and the implementation is publicly available.\footnote{\begin{rev} https://github.com/Robertocrotti98/hylog-hybridizer
    \end{rev}}
\end{enumerate}

Together, Hylog and the new Inputlog versions represent meaningful advances in keylogging for non-alphabetic scripts and contribute to more inclusive research on multilingual, digital text production. Extending Hylog to additional contexts will require addressing several challenges. For different IMEs (e.g., Sogou Pinyin, Baidu Pinyin), the pattern-matching rules \begin{revv}will\end{revv}
need adjustment to accommodate IME-specific behaviors, though the core architecture should remain applicable since all pinyin IMEs share fundamental features (syllable segmentation, candidate selection).

For other non-alphabetic scripts, the challenges vary: Japanese requires handling multiple scripts (hiragana, katakana, kanji) and input methods (romaji, kana input), necessitating additional pattern rules; Korean Hangul presents different complexities due to its alphabetic nature with syllabic composition; Arabic requires right-to-left handling. Each represents an important direction for future work, and we welcome collaboration with researchers working in these writing systems. Our open-source release aims to facilitate such extensions.

Fine-grained analyses of text production depend on ecological keylogging, yet traditional tools remain inadequate for non-alphabetic scripts that rely on Input Method Editors to convert keystrokes into characters. Keyloggers capture only motor actions, while text loggers capture only visible output. Hylog bridges this gap through a hybrid approach that unites both streams.

\begin{rev}In summary, the core technical contributions of Hylog\end{rev} are:
\begin{enumerate*}[label=(\roman*)]

\item two ecological text logger plug-ins for Microsoft Word and Google Chrome, respectively, that efficiently record only the portions of text modified at each step;

\item a hybridizer module that merges keystroke and text traces into a synchronized dual output, preserving how IME operations transform alphabetical input into Chinese characters;

\item an analyzer that performs IKI analysis for simplified Chinese by leveraging the structured information in the dual trace.
\end{enumerate*}

\begin{rev}In the paper we reported\end{rev} a short translation task from English into simplified Chinese \begin{rev}to illustrate\end{rev} Hylog's ability to produce rich and accurate interkeystroke interval (IKI) measures. Although still a prototype, it successfully and systematically identified \begin{rev}the large majorioty of\end{rev}
IME-related events and yielded reliable IKI analyses for Chinese text production.
Future work will extend beyond the controlled scenario described in our empirical evaluation \begin{rev}by considering user groups with different
\begin{revv}
typing behaviors,
\end{revv}
levels of proficiency and familiarity with the target language\end{rev}.
\begin{revv}
The implementation of Hylog we presented in this paper is limited to Simplfied Chinese and Microsoft IME, but its core architecture and the underlying\end{revv}
hybridization approach\begin{revv}
are\end{revv}
not tied to a specific IME or analysis type and should be \begin{revv}easily\end{revv}
adapted to other languages, pending future testing\begin{rev} and experimentation with a larger number of informants\end{rev}. Moreover, the plug-in architecture enables ecological text logging across a wide range of applications, provided suitable APIs are available.

In conclusion, this paper has introduced and validated a hybrid logging system that provides a viable method for studying text production in non-alphabetic scripts. By making the system and its architecture open-source, we hope to lower the barrier for researchers to conduct quantitative, ecological studies of writing and translation in languages like Chinese. Future work, both by our team and by the wider community, can use this tool to investigate the cognitive dynamics of IME typing, compare processes across L1 and L2 writers at scale, and ultimately contribute to a more inclusive study of digital, multilingual text production.

%% file: bibliography.bib
@article{vandermeulen2024,
 title={Getting a grip on the writing process:(Effective) approaches to write argumentative and narrative texts in L1 and L2},
 author={Vandermeulen, Nina and Lindgren, Eva and Waldmann, Christian and Levlin, Maria},
 journal={Journal of Second Language Writing},
 volume={65},
 pages={101113},
 year={2024},
 publisher={Elsevier}
}

@article{xu2014,
author = {Xu, Cuiqin and Ding, Yanren},
year = {2014},
month = {10},
pages = {80-96},
title = {An exploratory study of pauses in computer-assisted EFL writing},
volume = {18},
number  = {3},
journal = {Language Learning \& Technology},
publisher={University of Hawaii National Foreign Language Resource Center},
doi = {10.64152/10125/44385}
}

@article{raido2023,
 title = {Changes in web search query behavior of English-to-Chinese translation trainees},
 author = {{Enr{\'\i}quez Ra{\'\i}do}, Vanessa and Cai, Yuxing},
 journal = {Ampersand},
 volume = {11},
 pages = {100137},
 year = {2023},
 publisher = {Elsevier}
}

@article{leijten2019,
 title={MAPPING MASTER’S STUDENTS’USE OF EXTERNAL SOURCES IN SOURCE-BASED WRITING IN L1 AND L2},
 author={Leijten, Mari{\"e}lle and Van Waes, Luuk and Schrijver, Iris and Bernolet, Sarah and Vangehuchten, Lieve},
 journal={Studies in Second Language Acquisition},
 volume={41},
 number={3},
 pages={555--582},
 year={2019},
 publisher={Cambridge University Press}
}

@article{leijten2013,
 title={Keystroke logging in writing research: Using Inputlog to analyze and visualize writing processes},
 author={Leijten, Mari{\"e}lle and Van Waes, Luuk},
 journal={Written Communication},
 volume={30},
 number={3},
 pages={358--392},
 year={2013},
 publisher={Sage Publications Sage CA: Los Angeles, CA}
}

@article{lu2021,
 title={Revising in a non-alphabetic language: The multi-dimensional and dynamic nature of online revisions in Chinese as a second language},
 author={Lu, Xiaojun and R{\'e}v{\'e}sz, Andrea},
 journal={System},
 volume={100},
 pages={102544},
 year={2021},
 publisher={Elsevier}
}

@article{van2009,
 title={Keystroke logging in writing research: Observing writing processes with Inputlog},
 author={Van Waes, Luuk and Leijten, Mari{\"e}lle and Van Weijen, Daphne},
 journal={German as a foreign language},
 volume={2},
 pages={41--64},
 year={2009}
}

@article{couto2017,
 title={What does a translator do when not writing?},
 author={Couto Vale, Daniel},
 journal={Empirical modelling of translation and interpreting},
 volume={7},
 pages={209},
 year={2017},
 publisher={Language Science Press Berlin}
}

@article{rosa2018,
 title={Pauses by student and professional translators in translation process},
 author={Rosa, Rusdi Noor and Sinar, T Silvana and Ibrahim-Bell, Zubaidah and Setia, Eddy},
 journal={International Journal of Comparative Literature and Translation Studies},
 volume={6},
 number={1},
 pages={18--28},
 year={2018}
}

@article{munoz2018,
 author = {{Mu{\~n}oz Mart{\'\i}n}, Ricardo and Cardona Guerra, Jos{\'e} M.},
 year = {2018},
 month = {10},
 pages = {1--27},
 title = {Translating in fits and starts: pause thresholds and roles in the research of translation processes},
 volume = {27},
 journal = {Perspectives},
 doi = {10.1080/0907676X.2018.1531897}
}

@article{kasprowski2022,
AUTHOR = {Kasprowski, Pawel and Borowska, Zaneta and Harezlak, Katarzyna},
TITLE = {Biometric Identification Based on Keystroke Dynamics},
JOURNAL = {Sensors},
VOLUME = {22},
YEAR = {2022},
NUMBER = {9},
ARTICLE-NUMBER = {3158},
URL = {https://www.mdpi.com/1424-8220/22/9/3158},
PubMedID = {35590848},
ISSN = {1424-8220},
DOI = {10.3390/s22093158}
}

@article{acien2022,
 author = {{Aci{\'e}n Ayala}, Alejandro and {Morales Moreno}, Aythami and Monaco, John V. and {Vera Rodr{\'\i}guez}, Rub{\'e}n and {Fi{\'e}rrez Aguilar}, Juli{\'a}n},
 title = {{TypeNet}: Deep Learning Keystroke Biometrics},
 journal = {IEEE Transactions on Biometrics, Behavior, and Identity Science},
 year = {2022},
 volume = {4},
 number = {1},
 pages = {57--70},
 doi = {10.1109/TBIOM.2021.3112540}
}

@article{yang2021, 
title={TKCA: A timely keystroke-based continuous user authentication with short keystroke sequence in uncontrolled settings - cybersecurity},
url={https://cybersecurity.springeropen.com/articles/10.1186/s42400-021-00075-9#citeas},
journal={SpringerOpen},
publisher={Springer Singapore},
author={Yang, Lulu and Li, Chen and You, Ruibang and Tu, Bibo and Li, Linghui},
year={2021},
month={5}
}

@article{teh2013,
author = {Teh, Pin Shen and Teoh, Andrew and Yue, Shigang},
year = {2013},
month = {11},
pages = {408280},
title = {A Survey of Keystroke Dynamics Biometrics},
volume = {2013},
journal = {TheScientificWorldJournal},
doi = {10.1155/2013/408280}
}

@article{qassem2024,
 author = {Mutahar Qassem and Buthainah M. Al Thowaini},
 title = {Translation processes and products in L1-to-L2 and L2-to-L1 translations: Insights from keylogging data},
 journal = {Education and Information Technologies},
 year = {2024},
 volume = {29},
 number = {},
 pages = {21789--21809},
 doi = {10.1007/s10639-024-12689-w},
 publisher = {Springer},
 url = {https://doi.org/10.1007/s10639-024-12689-w}
}

@article{qassem2023,
 author = {Qassem, Mohammed and Al Thowaini, Bandar M.},
 title = {Cognitive Processes and Translation Quality: Evidence from Keystroke-Logging Software},
 journal = {Journal of Psycholinguistic Research},
 volume = {52},
 number = {5},
 pages = {1589--1604},
 year = {2023},
 doi = {10.1007/s10936-023-09964-1},
 publisher = {Springer}
}

@article{wang2025,
 author = {Wang, Feng and Xu, Qian},
 title = {Processing of Translation Units by Student and Semi-Professional Translators in Translating Texts with Different Levels of Translation Difficulty},
 journal = {PLOS ONE},
 volume = {20},
 number = {4},
 pages = {e0320809},
 year = {2025},
 doi = {10.1371/journal.pone.0320809},
 publisher = {Public Library of Science}
}

@article{dragsted2013, title={Towards a classification of translator profiles based on eye-tracking and keylogging data}, volume={5}, url={https://www.jowr.org/jowr/article/view/693}, DOI={10.17239/jowr-2013.05.01.6}, number={1}, journal={Journal of Writing Research}, author={Dragsted, Barbara and Carl, Michael}, year={2013}, month={6}, pages={133–158} }

@article{wang02092024,
author = {Yu Wang and Ali Jalalian Daghigh},
title = {Effect of text type on translation effort in human translation and neural machine translation post-editing processes: evidence from eye-tracking and keyboard-logging},
journal = {Perspectives},
volume = {32},
number = {5},
pages = {961--976},
year = {2024},
publisher = {Routledge},
doi = {10.1080/0907676X.2023.2219850},
URL = {https://doi.org/10.1080/0907676X.2023.2219850},
eprint = {https://doi.org/10.1080/0907676X.2023.2219850}
}

@article{sun2021, place={Antwerp, Belgium}, title={Effects of thinking aloud on cognitive effort in translation}, volume={19}, url={https://lans-tts.uantwerpen.be/index.php/LANS-TTS/article/view/556}, DOI={10.52034/lanstts.v19i0.556}, journal={Linguistica Antverpiensia, New Series – Themes in Translation Studies}, author={Sun, Sanjun and Li, Tian and Zhou, Xiaoyan}, year={2021}, month={1} }

@article{heilmann2021,
title = {Animacy and agentivity of Subject Themes in English-German translation},
journal = {Lingua},
volume = {261},
pages = {102813},
year = {2021},
note = {Dynamicity and Contrast in Systemic Functional Linguistics},
issn = {0024-3841},
doi = {https://doi.org/10.1016/j.lingua.2020.102813},
url = {https://www.sciencedirect.com/science/article/pii/S0024384120300206},
author = {Arndt Heilmann and Tatiana Serbina and Jonas Freiwald and Stella Neumann}
}

@article{sinulingga2023,
author = {Sinulingga, Amsaldi and Sofyan, Rudy and Mono, Umar},
year = {2023},
month = {03},
pages = {213-226},
title = {Time management and translation method in translating a scientific article: A case study on a professional translator},
volume = {8},
journal = {JOALL (Journal of Applied Linguistics and Literature)},
doi = {10.33369/joall.v8i1.22250}
}

@article{screen2016,
author = {Screen, Ben},
year = {2016},
month = {04},
pages = {1-18},
title = {What does Translation Memory do to translation? The effect of Translation Memory output on specific aspects of the translation process},
volume = {8},
journal = {Translation and Interpreting},
doi = {10.12807/ti.108201.2016.a01}
}

@article{miller2000,
author = {Kristyan Spelman Miller},
title ={Academic writers on-line: investigating pausing in the production of text},
journal = {Language Teaching Research},
volume = {4},
number = {2},
pages = {123-148},
year = {2000},
doi = {10.1177/136216880000400203},
URL = {https://doi.org/10.1177/136216880000400203},
eprint = {https://doi.org/10.1177/136216880000400203}
}

@article{baaijen2012,
author = {Veerle M. Baaijen and David Galbraith and Kees de Glopper},
title ={Keystroke Analysis: Reflections on Procedures and Measures},
journal = {Written Communication},
volume = {29},
number = {3},
pages = {246-277},
year = {2012},
doi = {10.1177/0741088312451108},
URL = {https://doi.org/10.1177/0741088312451108},
eprint = {https://doi.org/10.1177/0741088312451108}
}

@article{wengelin2006,
author = {Wengelin, {\AA}sa},
year = {2006},
month = {01},
pages = {107-130},
title = {Examining pauses in writing: Theories, methods and empirical data},
journal = {Computer Key-stroke Logging and Writing: Methods and Applications}
}

@article{ivaska2025, title={Pauses during a writing process in two typologically different languages}, volume={16}, url={https://www.jowr.org/jowr/article/view/1276}, DOI={10.17239/jowr-2025.16.03.03}, number={3}, journal={Journal of Writing Research}, author={Ivaska, Ilmari and Toropainen, Outi and Lahtinen, Sinikka}, year={2025}, month={2}, pages={407–433} }

@article{barkaoui2019, title={WHAT CAN L2 WRITERS’ PAUSING BEHAVIOR TELL US ABOUT THEIR L2 WRITING PROCESSES?}, volume={41}, DOI={10.1017/S027226311900010X}, number={3}, journal={Studies in Second Language Acquisition}, author={Barkaoui, Khaled}, year={2019}, pages={529–554}}

@conference{scriptlog,
title = "ScriptLog - an experimental keystroke logging tool",
author = "Victoria Johansson and Johan Frid and {\AA}sa Wengelin",
year = "2018",
language = "English",
booktitle = "ELN. 1st Literacy Summit, ELN ; Conference date: 01-11-2018 Through 03-11-2018",
}

@conference{carl2012,
 title = "{Translog-{{II}}}: A Program for Recording User Activity Data for Empirical Translation Process Research",
 author = "Michael Carl",
 year = "2012",
 language = "English",
 booktitle = "Proceedings of the 8th International Conference on Language Resources and Evaluation (LREC 2012)",
 note = "The 8th International Conference on Language Resources and Evaluation. LREC 2012 ; Conference date: 21-05-2012 Through 27-05-2012",
 url = "http://www.lrec-conf.org/lrec2012/?-Home-",
}

@inproceedings{lacruz2012,
 title={Average pause ratio as an indicator of cognitive effort in post-editing: A case study},
 author={Lacruz, Isabel and Shreve, Gregory M and Angelone, Erik},
 booktitle={Workshop on post-editing technology and practice},
 year={2012}
}

@inproceedings{tseng2005,
 title={A Conditional Random Field Word Segmenter for Sighan Bakeoff 2005},
 author={Tseng, Huihsin and Chang, Pi-Chuan and Andrew, Galen and Jurafsky, Dan and Manning, Christopher D.},
 booktitle={Proceedings of the Fourth SIGHAN Workshop on Chinese Language Processing},
 pages={168--171},
 year={2005}
}

@inproceedings{wang2001,
 author = {Jingtao Wang and Shumin Zhai and Hui Su},
 title = {Chinese Input with Keyboard and Eye-Tracking: An Anatomical Study},
 booktitle = {Proceedings of the 2001 Conference on Human Factors in Computing Systems (CHI '01)},
 year = {2001},
 pages = {349--356},
 publisher = {ACM},
 doi = {10.1145/365024.365298}
}

@inproceedings{Murphy2017,
  author={Murphy, Christopher and Huang, Jiaju and Hou, Daqing and Schuckers, Stephanie},
  booktitle={2017 IEEE International Joint Conference on Biometrics (IJCB)}, 
  title={Shared dataset on natural human-computer interaction to support continuous authentication research}, 
  year={2017},
  volume={},
  number={},
  pages={525-530},
  doi={10.1109/BTAS.2017.8272738}}

@inproceedings{Epp2011,
author = {Epp, Clayton and Lippold, Michael and Mandryk, Regan L.},
title = {Identifying emotional states using keystroke dynamics},
year = {2011},
isbn = {9781450302289},
publisher = {Association for Computing Machinery},
address = {New York, NY, USA},
url = {https://doi.org/10.1145/1978942.1979046},
doi = {10.1145/1978942.1979046},
booktitle = {Proceedings of the SIGCHI Conference on Human Factors in Computing Systems},
pages = {715–724},
numpages = {10},
location = {Vancouver, BC, Canada},
series = {CHI '11}
}

@inproceedings{Vural2014,
  author={Vural, Esra and Huang, Jiaju and Hou, Daqing and Schuckers, Stephanie},
  booktitle={IEEE International Joint Conference on Biometrics}, 
  title={Shared research dataset to support development of keystroke authentication}, 
  year={2014},
  volume={},
  number={},
  pages={1-8},
  doi={10.1109/BTAS.2014.6996259}}

@misc{genographixlog,
 title = {GenoGraphiX-Log project website},
 author = {Caporossi, Gilles and Leblay, Christophe and Usoof, Hakim},
 year = 2023,
 note = {Accessed: February 2025},
 howpublished = {\url{https://www.ggxlog.net}}
}

@misc{cywrite,
 title = {CyWrite GitHub page},
 author = {Evgeny, Chukharev},
 year = 2020,
 note = {Accessed: February 2025},
 howpublished = {\url{https://github.com/chukharev/cywrite}}
}

@misc{vsto,
  title = {Get started programming VSTO Add-ins},
  author = {{Microsoft®}},
  year = {2024},
  howpublished = {\url{https://learn.microsoft.com/en-us/visualstudio/vsto/getting-started-programming-vsto-add-ins?view=vs-2022}},
  note = {Last updated: 2024-03-11. Accessed: March 2025}
}

@incollection{dasilva2017,
  author       = {Lourenço da Silva, Igor A. and Alves, Fabio and Schmaltz, Márcia and Pagano, Adriana and Wong, Derek and Chao, Lidia and Leal, Ana Luísa V. and Quaresma, Paulo and Garcia, Caio and da Silva, Gabriel Eduardo},
  year         = {2017},
  title        = {Translation, post-editing and directionality: A study of effort in the Chinese-Portuguese language pair},
  editor       = {Jakobsen, Arnt Lykke and Mesa-Lao, Bartolomé},
  booktitle    = {Translation in Transition: Between Cognition, Computing and Technology},
  publisher    = {John Benjamins Publishing Company},
  pages        = {107--134},
  doi          = {10.1075/btl.133.04lou},
  isbn         = {9789027265371},
  url          = {https://doi.org/10.1075/btl.133.04lou},
  lastchecked  = {2025-11-20}
}

@incollection{wengelin2023,
 title={Investigating writing processes with keystroke logging},
 author={Wengelin, {\AA}sa and Johansson, Victoria},
 booktitle={Digital Writing Technologies in Higher Education: Theory, Research, and Practice},
 pages={405--420},
 year={2023},
 publisher={Springer}
}

@inbook{munoz2021,
author = {Muñoz Martín, Ricardo and Apfelthaler, Matthias},
year = {2021},
month = {11},
pages = {19-45},
publisher = {Springer},
title = {Spillover Effects in Task-Segment Switching: A study of translation subtasks as behavioral categories within the Task Segment Framework},
isbn = {978-981-16-2069-0},
doi = {10.1007/978-981-16-2070-6_2},
chapter = {2}
}

@phdthesis{sjrup2013,
address = {Frederiksberg},
author = {Annette Camilla Sj\o{}rup},
copyright = {https://creativecommons.org/licenses/by-nc-nd/3.0/},
school = {Copenhagen Business School (CBS)},
isbn = {9788792977496},
language = {eng},
note = {hdl:10398/8698},
number = {18.2013},
publisher = {Copenhagen Business School (CBS)},
series = {PhD Series},
title = {Cognitive effort in metaphor translation: An eye-tracking and key-logging study},
url = {https://hdl.handle.net/10419/208853},
year = {2013}
}

@phdthesis{lu2020,
 author = {Xiaojun Lu},
 title = {Writing in a non-alphabetic language using a keyboard: behaviours, cognitive activities and text quality},
 school = {University College London},
 year = {2020},
 type = {PhD thesis},
 note = {Green open access via UCL Discovery},
 url = {https://discovery.ucl.ac.uk/id/eprint/10106308/}
}

@phdthesis{du2024,
author = {Du, Zhiqiang},
year = {2024},
school = {University of Bologna},
month = {06},
pages = {},
title = {Bridging the gap: exploring the cognitive impact of InterpretBank},
doi = {10.48676/unibo/amsdottorato/11584}
}

@inproceedings{Dhakal2018,
 author = {Dhakal, Vivek and Feit, Anna Maria and Kristensson, Per Ola and Oulasvirta, Antti},
 title = {Observations on Typing from 136 Million Keystrokes},
 booktitle = {Proceedings of the 2018 CHI Conference on Human Factors in Computing Systems (CHI '18)},
 year = {2018},
 publisher = {Association for Computing Machinery},
 address = {New York, NY, USA},
 doi = {10.1145/3173574.3174220},
 url = {https://doi.org/10.1145/3173574.3174220}
}

@article{Immonen2011,
 author = {Immonen, Sini},
 title = {Unravelling the Processing Units of Translation},
 journal = {Across Languages and Cultures},
 year = {2011},
 volume = {12},
 number = {2},
 pages = {235--257},
 doi = {10.1556/Acr.12.2011.2.6}
}

@article{ImmonenMakisalo2010,
 author = {Immonen, Sini and M{\"a}kisalo, Jukka},
 title = {Pauses Reflecting the Processing of Syntactic Units in Monolingual Text Production and Translation},
 journal = {Hermes -- Journal of Language and Communication in Business},
 year = {2010},
 volume = {44},
 pages = {45--61},
 doi = {10.7146/hjlcb.v23i44.97266}
}

@incollection{Munoz2025,
 author = {Mu{\~n}oz Mart{\'i}n, Ricardo and Sun, Sanjun and Du, Zhiqiang and Puerini, Sara},
 title = {Keylogging},
 booktitle = {Research Methods in Cognitive Translation and Interpreting Studies},
 publisher = {John Benjamins},
 address = {Amsterdam},
 year = {2025},
 pages = {157--182}
}

@article{TianCushing2025,
  author = {Tian, Yu and Cushing, Sara T.},
  title = {Exploring the Application of Keystroke Logging Techniques to Research in Second Language (L2) Writing},
  journal = {Research Methods in Applied Linguistics},
  year = {2025},
  volume = {4},
  number = {1},
  pages = {100179},
  doi = {10.1016/j.rmal.2024.100179}
}

@article{Sun2016,
  title={Shared keystroke dataset for continuous authentication},
  author={Yan Lindsay Sun and Hayreddin Çeker and Shambhu J. Upadhyaya},
  journal={2016 IEEE International Workshop on Information Forensics and Security (WIFS)},
  year={2016},
  pages={1-6},
  url={https://api.semanticscholar.org/CorpusID:15959476}
}

@article{Katerina2018,
title = {Mouse behavioral patterns and keystroke dynamics in End-User Development: What can they tell us about users’ behavioral attributes?},
journal = {Computers in Human Behavior},
volume = {83},
pages = {288-305},
year = {2018},
issn = {0747-5632},
doi = {https://doi.org/10.1016/j.chb.2018.02.012},
url = {https://www.sciencedirect.com/science/article/pii/S0747563218300700},
author = {Tzafilkou Katerina and Protogeros Nicolaos}
}
